\documentclass{article}

\usepackage[nonatbib, preprint]{neurips_2024}

\usepackage[utf8]{inputenc} % allow utf-8 input
\usepackage[T1]{fontenc}    % use 8-bit T1 fonts
\usepackage{hyperref}       % hyperlinks
\usepackage{url}            % simple URL typesetting
\usepackage{booktabs}       % professional-quality tables
\usepackage{amsfonts}       % blackboard math symbols
\usepackage{nicefrac}       % compact symbols for 1/2, etc.
\usepackage{microtype}      % microtypography
\usepackage{subfigure}

\usepackage{epsfig}
\usepackage{graphicx}
\usepackage{amsmath}
\usepackage{amssymb}
\usepackage{amsthm}
\usepackage{adjustbox}
\usepackage{caption}
\usepackage{verbatim}

\usepackage{algorithm}
\usepackage{algorithmic}
\usepackage{enumitem}

\newcommand{\mypar}[1]{\vspace{3pt}\noindent\textbf{#1~}}

\usepackage{colortbl}

\usepackage{pifont}

\usepackage{graphicx}
\usepackage{amsmath}
\usepackage{amssymb}
\usepackage{booktabs}
\usepackage{multirow}
\usepackage{enumitem}

\usepackage{array}      % for array and tabular features
\usepackage{graphicx}   % for \resizebox or \scalebox

\usepackage{graphicx}
\usepackage{amsmath}
\usepackage{amssymb}
\usepackage{booktabs}
\usepackage{multirow}
\usepackage{enumitem}
\usepackage[table]{xcolor}

\newcommand{\ccol}{\cellcolor{blue!8}}

\usepackage{glossaries}

\newacronym{tta}{TTA}{Test-Time Adaptation}
\newacronym{fps}{FPS}{Farthest Point Sampling}
\newacronym{knn}{KNN}{K-Nearest Neighbors}
\newacronym{maes}{MAEs}{Masked Autoencoders}
\newacronym{mae}{MAEs}{Masked Autoencoder}
\newacronym{ssl}{SSL}{Self-Supervised Learning}
\newacronym{ttt}{TTT}{Test-Time Training}
\newacronym{wa}{WA}{Weight Averaging}
\newacronym{ln}{LN}{Layer Normalization}
\newacronym{bn}{BN}{Batch Normalization}

\title{Test-Time Adaptation in Point Clouds: Leveraging Sampling Variation with Weight Averaging}

\author{Ali Bahri\thanks{Correspondence to \href{mailto:mehrdad.noori.1@ens.etsmtl.ca}{ali.bahri.1@ens.etsmtl.ca}} \And Moslem Yazdanpanah \And Mehrdad Noori \And Sahar Dastani \And Milad Cheraghalikhani \And David Osowiech \AND Farzad Beizaee \And Gustavo adolfo.vargas-hakim \And Ismail Ben Ayed\And Christian Desrosiers\AND \
LIVIA, ÉTS Montréal, Canada \\
International Laboratory on Learning Systems (ILLS), \\
McGILL - ETS - MILA - CNRS - Université Paris-Saclay - CentraleSupélec, Canada}

\newsavebox\CBox

\begin{document}

\maketitle

\begin{abstract}
Test-Time Adaptation (TTA) addresses distribution shifts during testing by adapting a pretrained model without access to source data. In this work, we propose a novel TTA approach for 3D point cloud classification, combining sampling variation with weight averaging. Our method leverages Farthest Point Sampling (FPS) and K-Nearest Neighbors (KNN) to create multiple point cloud representations, adapting the model for each variation using the TENT algorithm. The final model parameters are obtained by averaging the adapted weights, leading to improved robustness against distribution shifts. Extensive experiments on ModelNet40-C, ShapeNet-C, and ScanObjectNN-C datasets, with different backbones (Point-MAE, PointNet, DGCNN), demonstrate that our approach consistently outperforms existing methods while maintaining minimal resource overhead. The proposed method effectively enhances model generalization and stability in challenging real-world conditions. The implementation is available at: \url{https://github.com/AliBahri94/SVWA_TTA.git}.
\end{abstract}

\section{Introduction}
\label{sec:intro}

Deep neural networks have recently demonstrated impressive capabilities in classifying 3D point clouds \cite{qi2017pointnet, qi2017pointnet++, wang2019dynamic, pang2022masked, zhang2022point, zhang2023learning}. However, this success typically relies on the assumption that the test data is drawn from the same distribution as the training data. In real-world applications, this assumption is often invalid. 
%For example, a point cloud captured by a drone's sensor can be distorted due to equipment malfunction or challenging environmental factors like weather conditions or obstacles. 
When the test distribution (\emph{target}) differs from the training distribution (\emph{source}), the challenge of distribution shifts arises. In 3D data, such differences can vary widely, as they may be caused by various factors including the type sensor (e.g., RGB-D camera or Lidar), conditions of the environment (e.g., low light for RGB-D camera), and occlusions. This make it impractical to pretrain the network for every possible shift encountered during testing. It is thus essential to develop methods that can adapt to these distribution changes in real-time, and without supervision, during the test phase.

By addressing a more realistic setting where distribution shifts can also occur after training, \gls{tta} recently became a focal point for researchers in machine learning and computer vision \cite{gandelsman2022test, gao2023back, iwasawa2021test, lim2023ttn, yeo2023rapid, schneider2020improving}. \gls{tta} uses unlabeled test data to adapt a source-pretrained model to distribution shifts occurring in the testing phase. In this paper, we consider the fully-\gls{tta} setting where the model is pretrained on source data in a standard supervised manner, without any additional mechanism for adaptation, and the model is only adapted in testing. This setting contrast with Test-Time Training (TTT), where specialized strategies for adaptation are incorporated during source pretraining. 

%\gls{tta} is particularly important because it allows models to rapidly adapt to distribution shifts in real-time during testing, ensuring accurate final predictions and enhancing their robustness and performance in unpredictable or dynamic environments.

%For instance, imagine an autonomous drone equipped with a camera-based object detection system designed to recognize obstacles and navigate through various terrains. Although this system may have been trained in ideal weather conditions, it must rapidly adapt to new and challenging environments during test time, such as encountering heavy fog or strong winds. This scenario emphasizes the crucial need for \gls{tta} to enable the drone to adapt effectively to these dynamic conditions without requiring extensive retraining.

% \gls{tta} uses unlabeled test data to adapt a classifier in real-time to shifts in data distributions during the testing phase. The pretrained model on the source data is typically trained in a fully supervised manner for tasks such as classification, without employing any additional techniques, ensuring that adaptation occurs solely during testing. Recently, various \gls{tta} methods have been introduced in the 2D image domain. Key strategies include regularizing the classifier on test data using objective functions based on the entropy of its predictions \cite{liang2020we, zhang2022memo}, and updating batch normalization statistics to align with the test data distribution \cite{mirza2022norm}. 

In recent years, various \gls{tta} methods have been introduced in the 2D image domain. Key strategies include regularizing the classifier on test data using objective functions based on the prediction entropy \cite{liang2020we, zhang2022memo}, or updating batch normalization statistics to align with the test data distribution \cite{mirza2022norm}. In the context of 3D point cloud classification, \gls{tta} is a relatively new and emerging field, with only two approaches proposed for this task: MATE \cite{mirza2023mate} and BFTT3D \cite{wang2024backpropagation}. However, MATE \cite{mirza2023mate} can technically be categorized as Test-Time Training rather than \gls{tta}, as it involves using a masked autoencoder during the source pretraining phase. On the other hand, BFTT3D \cite{wang2024backpropagation} employs a set of source prototypes to adapt to new target domains. While this prototype memory maintains privacy, it does not fully align with the core principle of \gls{tta} which aims to avoid reliance on source data during adaptation. %This method's dependence on prototype memory indicates a compromise between needing some knowledge from the source domain and striving for complete independence during adaptation.

In this paper, we propose the first fully-\gls{tta} strategy for 3D point cloud classification. Our approach is inspired by the concept of seeking flat minima via weight averaging as highlighted in the SWA \cite{izmailov2018averaging} and SWAD \cite{cha2021swad} papers. Focusing on flat minima, our method aims to enhance model robustness against distribution shifts, which are common in real-world scenarios. A key innovation in our approach is the use of sampling variation to drive the adaptation. Specifically, we employ \gls{fps} and \gls{knn} to generate multiple variations of sampled points within the point cloud, thereby introducing controlled stochasticity during adaptation. By combining the weights obtained using differently-sampled point clouds, the model is steered away from sharp minima which are more prone to overfitting and less robust to distribution shifts.

The iterative adaptation process of our method, which is guided by the prediction entropy minimization strategy of TENT \cite{wang2020tent}, ensures that each variation in the sampling contributes to a broader exploration of the loss landscape. By saving the model weights after each adaptation, and subsequently averaging them, we converge to a flatter and more stable region in the loss landscape. This weight averaging technique, inspired by the SWAD approach, mitigates the impact of noise and outliers within individual samples, leading to a more robust and generalizable model. 

%The resulting model parameters, being a product of multiple adaptation scenarios, exhibit enhanced resilience to distribution shifts, thereby improving overall performance on corrupted datasets during the testing phase.

We outline the main contributions of our work as follows:
\begin{itemize}[itemsep=-3pt,topsep=0pt]
    \item \textbf{Novelty}: Addressing the lack of studies in this field, we introduce the first fully-\gls{tta} method specifically designed for 3D point cloud classification. Our method proposes a novel strategy for this challenging task, which combines sampling variation and weight averaging at test time.
    \item \textbf{Robustness}: Our method, which achieves complete \gls{tta} without accessing any source data, demonstrates superior efficiency compared to leading approaches like TENT even with very small batch sizes.
    \item \textbf{State-of-art performance}: Through an extensive set of experiments involving three datasets modeling a broad range of corruptions and three different backbones for point cloud classification, we show that our method achieves state-of-art performance in most test cases.
\end{itemize}

\begin{figure*}[t]
    \centering
    \includegraphics[width=1\textwidth]{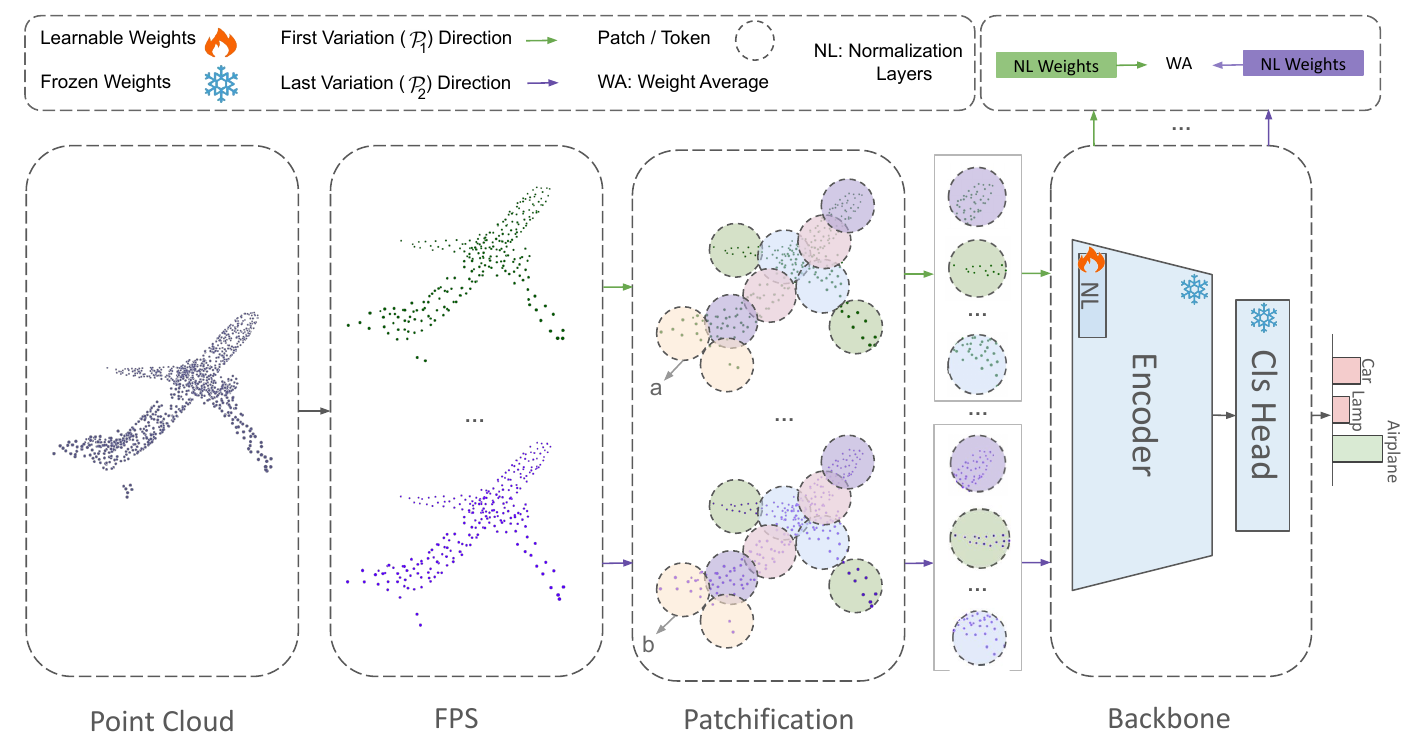}
    \caption{Overview of our 3D \gls{tta} methodology. First, \gls{fps} is applied to generate different samplings from the input point cloud. Patchification is then performed using \gls{fps} for patch centers and \gls{knn} to form patches (a and b). The Normalization Layer (NL) weights are adapted using the TENT algorithm for each sampling. Finally, weight averaging is applied across all adapted weights to enhance robustness and generalization.
        }
    \label{fig:method}
\end{figure*}

% \vspace{-10 pt}

\section{Related Work}
\label{sec:related_work}

\mypar{Test-Time Adaptation.} 
\gls{tta} tackles domain adaptation in the more realistic and challenging scenario where the target domain data is unlabeled and we have no access to source domain samples. The primary challenge of this task involves accurately estimating the target domain's distribution and indirectly comparing it to the source domain's characteristics.
A typical approach to reduce domain shift when source data is unavailable involves fine-tuning the model using an unsupervised loss based on the target distribution. The TTT algorithm \cite{sun2020test} enhances the model by updating its parameters in real-time through a self-supervised task applied to the test data. TENT \cite{wang2020tent} updates the trainable batch normalization parameters during testing by minimizing the entropy of the model's predictions. Source hypothesis transfer (SHOT) \cite{liang2020we} combines prediction entropy minimization with a diversity regularization prior (maximizing the entropy of the class marginal) to train a robust feature extractor from a pretrained source model. TTT++ \cite{liu2021ttt++} incorporates an additional self-supervised branch that utilizes contrastive learning within the source model to aid in adapting to the target domain. TTTFlow \cite{osowiechi2023tttflow} utilizes unsupervised normalizing flows as an alternative to self-supervision for the auxiliary task.

\mypar{Test-Time Point Cloud Adaptation.} 
The concept of \gls{tta}, initially tailored for 2D images, often faces challenges when applied to 3D data, necessitating specialized approaches. So far, very few works have studied this problem in the context of 3D point cloud data. One of the first \gls{ttt} approaches specifically designed for 3D data, MATE \cite{mirza2023mate}, employs a Masked Autoencoder (MAE) reconstruction objective to enhance the robustness of a point cloud classification network to distribution shifts in test data. The Continual Test-Time Domain Adaptation (CTDA) method \cite{wang2024continual} employs Dynamic Sample Selection (DSS) to handle noisy pseudo-labels while adapting a pretrained model to new target domains without accessing source data. This approach, enhancing model performance through dynamic thresholding and positive-negative learning, has proven effective in both the 2D image and 3D point cloud domains. 

The work in \cite{shin2022mm} presents a multi-modal extension of \gls{tta} for 3D semantic segmentation. The proposed method introduces Intra-modal Pseudolabel Generation (Intra-PG) to generate reliable pseudo labels within each modality and Inter-modal Pseudo-label Refinement (Inter-PR) to refine these labels across modalities. Hatem et al. \cite{hatem2023test} introduced a \gls{tta} technique for point cloud upsampling, leveraging meta-learning to improve model generalization during inference. In point cloud registration, Point-TTA \cite{hatem2023point} offers a \gls{tta} framework improving model generalization by adapting to each test instance through self-supervised auxiliary tasks. This method allows the model to handle unseen data distributions during testing without prior knowledge. Finally, BFTT3D \cite{wang2024backpropagation} introduces a backpropagation-free Test-Time Adaptation (TTA) method specifically designed for 3D data, addressing domain shifts with a two-stream architecture that maintains both source and target domain knowledge. However, this approach does not fully align with the core principle of TTA, which aims to avoid reliance on source data during adaptation. Zhang et al. \cite{zhang2022memo}  introduced MEMO, a method that enhances model robustness during test time by applying data augmentations to a single test input and adapting model parameters to minimize the entropy of the averaged output distribution across these augmentations.

\mypar{Weight Averaging.} 
Weight averaging has become a prominent technique for enhancing the generalization of deep neural networks during training. Stochastic Weight Averaging (SWA) \cite{izmailov2018averaging} improves model generalization by averaging weights from different training epochs, promoting smoother optimization and convergence to well-generalized solutions. Building on this, SWAD \cite{cha2021swad} refines the approach by densely sampling weights throughout the training process, further boosting generalization and robustness across tasks. Addressing limitations of traditional WA techniques that average weights post-training, Lookaround \cite{zhang2024lookaround} introduces a novel optimization strategy integrating diversity into the training process. This method iteratively alternates between two steps: the ``around step,'' which enhances network diversity by training multiple models on differently augmented data, and the ``average step'' which consolidates these models into a single network. %This approach leads to flatter minima and better generalization

%-------------------------------------------------------------------------
\section{Method}
\label{sec:method}

We propose a novel \acrfull{tta} strategy tailored for 3D point cloud classification, which focuses on enhancing model robustness against distribution shifts. Our method introduces a dual approach: first, we create diverse perspectives of the input data to address distributional changes by utilizing sampling variation; second, we integrate this variation with a weight averaging technique. This combination operates within a purely test-time framework, eliminating the need for any source data during the adaptation process. The overview of our \gls{tta} method for point cloud data is presented in Figure~\ref{fig:method}.

\subsection{Sampling Variation}

We start with a 3D point cloud $P \in \mathbb{R}^{N^p \times 3}$ consisting of $N^p$ points. To address the challenges posed by distribution shifts, we introduce a method that leverages sampling variation during test-time adaptation. This begins by creating patches from the point cloud, where each patch is defined by a center point $c_i$ and its neighboring points.
We employ \gls{fps} to select a set of center points $\{c_1, c_2, \dots, c_N\}$. \gls{fps} works by iteratively selecting the point in $P$ that is furthest from all previously selected centers:
\begin{equation}
c_i \, = \, \arg \max_{p \in P} \, \min_{c_j \in \{c_1, \dots, c_{i-1}\}} \, \| p - c_j \|_2
\label{eq:fps}
\end{equation}
Then, for each center $c_i$, we use \gls{knn} to find its neighboring points, forming a patch of neighboring points $P_i$ as follows:
\begin{equation}
P_i \, = \, \Bigl\{ p \in P \ | \ p \in \text{KNN}(c_i) \Bigr\}
\label{eq:knn}
\end{equation}
This process converts the point cloud into a collection of patches $\mathcal{P} \in \mathbb{R}^{N \times K \times 3}$, where $N$ is the number of patches and $K$ is the number of neighbors in each patch (including the center).

During test-time adaptation, we generate multiple versions of $\mathcal{P}$ by varying the selection of centers and neighbors forming $\{\mathcal{P}_1, \mathcal{P}_2, \dots, \mathcal{P}_V\}$ of size $N^V$. For Non-Transformer networks, we assume $K\!=\!1$, meaning that each patch consists of only the center point selected by \gls{fps} without including any neighbors.
%which can be described by introducing variations $ \delta P_i$ to the patch $ P_i$:
% \begin{equation}
% \mathcal{P}_{\text{patch}}^i \, = \, P_i + \delta P_i
% \label{eq:sampling_variation}
% \end{equation}

% \[
% \mathcal{P}_{\text{patch}}^i \, = \, \text{FPS}^i(P)
% \label{eq:sampling_variation}
% \]

% \begin{equation}
% \mathcal{P}_{\text{patch}}^j \, = \, \{ (P_j^1 + \delta P_j^1), (P_j^2 + \delta P_j^2), \dots, (P_j^K + \delta P_j^K) \}
% \label{eq:sampling_variation_transformer}
% \end{equation}

This variation in sampling creates different representations of the same underlying 3D structure and can be interpreted as transformations that slightly modify the local geometric structure of patches while maintaining the overall global structure of the point cloud. This helps the model generalize better by encouraging it to learn robust features that are less sensitive to such variations (e.g., $\mathrm{a}$ and $\mathrm{b}$ in Figure~\ref{fig:method}). 

For each $\mathcal{P}_v$, the model is adapted using the TENT \cite{wang2020tent} algorithm, where normalization layers' parameters $\gamma$ and $\beta$ are updated to minimize the entropy $\mathcal{H}$ of the model's output:
\begin{equation}
\gamma^*_i, \beta^*_i \, = \, \arg\min_{\gamma, \beta} \, \mathbb{E} \left[\mathcal{H}\left(f_{\theta}\left(\mathcal{P}_v\right)\right)\right].
\label{eq:tent}
\end{equation}

Here, the expectation $\mathbb{E}$ is taken over the distribution of sampling variations $\mathcal{P}_v$, and $\mathcal{H}$ represents the entropy of the model's predictions, which is minimized to encourage confident and stable predictions.

%The integration of sampling variation provides several critical advantages. By generating multiple configurations of the same point cloud, our method captures more robust geometric features, mitigating the effects of noise and sensor inaccuracies. 
%This approach aligns with the IAE framework \cite{yan2023implicit}, where reconstruction under different variations encourages the encoder to focus on stable geometric features rather than being misled by noise-induced imperfections.

%This process ensures that the model adapts to various local minima of the loss function landscape, thus avoiding overfitting to specific configurations of the input data. 
%By forcing the model to adapt to different configurations of the same point cloud, we ensure that it does not rely on any single viewpoint or sample. This reduction in reliance on specific data patterns enhances the model's generalization capabilities, making it more resilient to unseen distribution shifts. 
%The introduction of sampling variation is crucial for improving the model's ability to generalize to new environments during test-time adaptation, leading to better performance in real-world scenarios where distribution shifts are common.

\subsection{Integrating Sampling Variation with Weight Averaging}

For each variation \( \mathcal{P}_v \), the model is adapted using the TENT algorithm, as outlined in Equation (\ref{eq:tent}).
The key innovation here is to combine these adaptations using a refined weight averaging technique inspired by \cite{cha2021swad, zhang2024lookaround}, leading to a more stable and generalizable model.

The concept of \textbf{weight averaging} aims to identify a solution in the parameter space that resides within a \emph{flat region} of the loss landscape. Flat minima are characterized by a low loss that remains relatively constant under small variations of the model parameters. Solutions in such regions tend to generalize better to distribution shifts because the model's performance is less sensitive to minor variations or noise in the input data and model parameters.

% In our effort to maximize variation for the model, we employed \gls{fps} to introduce diverse sampling variations in the data. FPS induces significant variations in network performance, making it an effective tool for this purpose. While we experimented with other data augmentation techniques, they did not complement the weight averaging process as effectively as FPS. The details of these experiments and their outcomes are presented in the ablation studies section.

This integrating sampling variation with weight averaging is related to Robust Risk Minimization (RRM) \cite{cha2021swad}. In our method, we follow a similar principle in the context of test-time adaptation. In RRM, the goal is to minimize the worst-case empirical loss within a neighborhood of model parameters. This is formulated as
\begin{equation}
\hat{\mathcal{E}}_{\mathcal{D}}^{\gamma}(\theta) \, = \, \max_{\|\Delta\| \leq \gamma} \, \hat{\mathcal{E}}_{\mathcal{D}}(\theta + \Delta),
\label{eq:robust_empirical_loss}
\end{equation}
where $\gamma > 0$ defines the neighborhood around the model parameters $\theta$, and $\Delta$ represents small perturbations. %By minimizing this robust empirical loss, RRM identifies parameters in flat regions of the loss landscape where performance remains stable under small variations.

While we do not explicitly optimize this loss function, the concept of locating flat minima is analogous to our use of parameter averaging. By averaging the weights of models adapted from different \( \mathcal{P}_v \), we effectively find a solution in the weight space that resides in a flatter region of the loss landscape. This improves the model's generalization during test-time adaptation and enhances its robustness to distribution shifts.

In our approach, after adapting the model for each variation \( \mathcal{P}_v \), we store the adapted weights $\theta_v$. These weights are then averaged to obtain the final model weights $\theta_{\text{avg}}$:
\begin{equation}
\theta_{\text{avg}} \, = \, \frac{1}{N^V} \sum_{v=1}^{V} \theta_v
\label{eq:wa}
\end{equation}

This averaging process provides two key benefits:
\begin{enumerate}[itemsep=2pt,topsep=2pt]
    \item \textbf{Flat Minima:} Averaging the weights from different adapted models helps to locate a point in the weight space that is situated at the intersection of several flat regions. This approach reduces the model's sensitivity to specific input data configurations, thereby enhancing its robustness against distribution shifts.
    
    \item \textbf{Error Reduction:} The averaging process mitigates the influence of errors or noise that may be present in individual model adaptations. By smoothing out fluctuations in different sub-samples, the average model becomes more stable.
\end{enumerate}
% Moreover, the integration of sampling variation ensures that the averaged model is not biased toward any single configuration of the input data. Instead, it leverages diverse perspectives, making the final model more resilient to variations in the data distribution.

% By combining the benefits of sampling variation and weight averaging, our method achieves a significant improvement in robustness and generalization compared to traditional TTA methods. This dual approach allows the model to maintain high performance across different target domains, even in the absence of source data.
%-------------------------------------------------------------------------

\section{Experiments}
\label{sec:experiments}
In this section, we conduct a comprehensive evaluation of our proposed method across multiple 3D point cloud datasets, focusing on both Transformer-based and non-Transformer-based backbones. To thoroughly assess the robustness and generalization capabilities of our approach, we perform experiments on three benchmark datasets: \emph{ModelNet40-C}, \emph{ShapeNet-C}, and \emph{ScanObjectNN-C}. These datasets encompass a range of real-world challenges, including varying levels of corruption and noise, allowing us to demonstrate the effectiveness of our method in diverse and complex scenarios.

\subsection{Implementation Details}

% During the TTA phase, we utilized multiple backbones -- Point-MAE, PointNet, and DGCNN -- as the source models, each independently trained on its corresponding clean dataset to evaluate the robustness of our method. Pretrained backbones were adapted using our proposed approach. For all backbones, we used the AdamW optimizer with a learning rate of 0.001, consistent with the base learning rate of the TENT algorithm. All the settings and hyperparameters are provided in the Supplementary Material.
During the TTA phase, we utilized multiple backbones, Point-MAE, PointNet, and DGCNN, as the source models. Each one is independently trained on its corresponding clean dataset to evaluate the robustness of our method. Pretrained backbones were adapted using our proposed approach. For all backbones, we used the AdamW optimizer with a learning rate of 0.001, consistent with the base learning rate of the TENT algorithm. Detailed information on the settings and hyperparameters are provided in the Supplementary Material. Additionally, \textbf{Resource Overhead -- Time} and \textbf{Resource Overhead -- Memory} related to our method are discussed in detail in the Supplementary Material.

For the Transformer-based Point-MAE backbone, we explored two distinct adaptation strategies. In the first approach, only the \gls{bn} layers were adapted, following the standard TENT adaptation approach. In the second approach, we adapted both \gls{ln} and \acrfull{bn} layers. This configuration was then compared with the TENT algorithm, which similarly adapts both \gls{ln} and \gls{bn} layers. The results of both approaches are compared in detail in the Supplementary Material.

We only adapted the \gls{bn} layers for the non-transformer-based PointNet and DGCNN backbones since these architectures do not \gls{ln} layers. The input point cloud size was set to 1024 points for all experiments. All experiments were conducted using a single NVIDIA A6000 GPU, ensuring consistency across all tested configurations.

\subsection{Datasets}

\mypar{ModelNet-40C.} 
ModelNet-40C \cite{sun2022benchmarking} is a robustness benchmark for point cloud classification, designed to assess how well architectures can handle real-world distribution shifts. It introduces 15 common corruption types to the original ModelNet-40 test set, categorized into three groups: transformation, noise, and density. These corruptions simulate real-world issues, such as sensor faults and noise in LiDAR scans, providing a realistic challenge for evaluating model performance under varying conditions. %The dataset aims to test models' resilience to changes that are typical in practical applications.

\mypar{ShapeNet-C.}
ShapeNetCore-v2 \cite{chang2015shapenet} is a large-scale dataset used for point cloud classification, consisting of 51,127 shapes from 55 categories. It is divided into training (70\%), validation (10\%), and test (20\%) sets. 
To assess model robustness under real-world conditions, \cite{mirza2023mate} applies 15 different types of corruptions to the test set, similar to those in ModelNet-40-C. These corruptions were generated using an open-source implementation provided by \cite{sun2022benchmarking}. This modified version of the dataset is referred to as ShapeNet-C.

\mypar{ScanObjectNN-C.}
ScanObjectNN \cite{uy2019revisiting} is a real-world point cloud classification dataset consisting of 15 categories. It includes 2,309 samples for training and 581 samples for testing. To evaluate model robustness, \cite{mirza2023mate} introduces 15 distinct corruptions to the test set, following the methodology outlined in \cite{sun2022benchmarking}. This modified version of this dataset is referred to as ScanObjectNN-C.

\begin{table*}[b!]
\setlength\tabcolsep{4pt}
\centering
\resizebox{0.9\textwidth}{!}{
\small
\begin{tabular}{l|l|ccccccccccccccc|cc}
\toprule
&
Method & \rotatebox{60}{uni} & \rotatebox{60}{gauss} & \rotatebox{60}{backg} & \rotatebox{60}{impul} & \rotatebox{60}{upsam} & \rotatebox{60}{rbf} & \rotatebox{60}{rbf-inv} & \rotatebox{60}{den-dec} & \rotatebox{60}{dens-inc} & \rotatebox{60}{shear} & \rotatebox{60}{rot} & \rotatebox{60}{cut} & \rotatebox{60}{distort} & \rotatebox{60}{oclsion} & \rotatebox{60}{lidar} & \rotatebox{0}{Mean}\\
\midrule
\parbox[t]{4mm}{\multirow{9}{*}{\rotatebox[origin=c]{90}{Point-MAE}}} &
Source-Only & 66.6 & 59.1 & 7.2 & 31.8 & 74.6 & 67.7 & 69.8 & 59.3 & 75.1 & 74.4 & 38.0 & 53.7 & 70.0 & 38.6 & 23.4 & 53.9 \\
&
DUA* \cite{mirza2022norm} & 65.0 & 58.5 & 14.7 & 48.5 & 68.8 & 62.8 & 63.2 & 62.1 & 66.2 & 68.8 & 46.2 & 53.8 & 64.7 & 41.2 & 36.5 & 54.7 \\
&
TTT-Rot* \cite{sun2020test} & 61.3 & 58.3 & 34.5 & 48.9 & 66.7 & 63.6 & 63.9 & 59.8 & 68.6 & 55.2 & 27.3 & 54.6 & 64.0 & 40.0 & 29.1 & 53.0 \\
& 
T3A* \cite{iwasawa2021test} & 64.1 & 62.3 & 33.4 & 65.0 & 75.4 & 63.2 & 66.7 & 57.4 & 63.0 & 72.7 & 32.8 & 54.4 & 67.7 & 39.1 & 18.3 & 55.7 \\
&
SHOT \cite{liang2020we} &76.9 & 71.2&	21.4&	63.6&	79.0	&71.2	&73.3	&75.0	&79.6&	76.6&	55.7&	71.8&	71.2&	46.1&	44.7 & 65.1\\
&
MATE* \cite{mirza2023mate} & 75.0& 71.1& 27.5& 67.5& 78.7& 69.5& 72.0& 79.1& 84.5& 75.4& 44.4& 73.6& 72.9& 39.7& 34.2 & 64.3\\
& 
PL \cite{lee2013pseudo} &  81.9&	78.1	&25.4	&69.3&	77.0&	77.4	&78.4	&83.7&	86.8&81.4&	63.0&	82.3&	78.1	&52.3&	49.5&	71.0\\
\cmidrule{2-19}
& 
MEMO \cite{zhang2022memo} & 83.8&	81.0&	30.8	&68.7	& 83.6&	75.3&  77.3 &	83.0  &	85.3	& 74.6&	58.0 &	74.9	& 67.6 &	50.3 &	47.3&   69.4   \\
&
TENT \cite{wang2020tent} &  82.0	&78.6&	26.7&	71.0&	78.2&	78.4&	79.9&	84.6&	87.0&	82.7&	65.4&	83.5&	79.6&	52.3&	51.2& 72.1\\
&
\ccol Ours & \ccol \textbf{85.0}	& \ccol \textbf{83.9}	&\ccol \textbf{33.0}	&\ccol \textbf{74.6}	&\ccol \textbf{87.0}&	\ccol \textbf{80.9}&	\ccol \textbf{82.3}&	\ccol \textbf{85.1}	&\ccol \textbf{88.0}	&\ccol \textbf{82.7}	&\ccol \textbf{66.9}	&\ccol \textbf{84.0}&	\ccol \textbf{80.5}&	\ccol \textbf{56.2}&	\ccol \textbf{55.3}& \ccol \textbf{75.0}   \\
\midrule
\midrule
\parbox[t]{4mm}{\multirow{7}{*}{\rotatebox[origin=c]{90}{PointNet}}} & 
Source-Only &   43.2&	82.8&	4.1& 41.6	&43.8	&44.1	&44.6 &	84.3 &	86.2& 33.6	& 24.0 &	83.6 &	37.9&	22.6&	21.8& 46.5 \\
%TTT-Rot &  & 5.2& 	3.9& 	3.6	& 3.0& 	3.9& 	3.2	& 3.6& 	3.8	& 3.1& 	3.9& 	3.6& 	4.1& 	4.2& 	2.5& 	4.6& 	3.7  \\
& 
LAME \cite{boudiaf2022parameter} & 23.4 & 15.3 & 4.0 & 4.5 & 32.2 & 6.6 & 8.8 & 40.1 & 65.2 & 3.5 & 2.5 & 31.48 & 7.5 & 4.0 & 4.0 & 16.8 \\
&
SHOT \cite{liang2020we} &  71.8&	83.5&	11.9&	66.4	&71.2&	61.3&	61.6&	82.7&	83.7&	41.2	&29.1&	80.5&	49.0&	28.7&	28.7&  56.8     \\
&
DUA \cite{mirza2022norm} & 67.5 &	84.7	&8.3	&61.8&	67.1	&60.2&	60.7	&84.6&	86.3&	43.0 &	31.1 &	84.2	&51.5&	32.9&	35.6&	57.3 \\
&
PL \cite{lee2013pseudo} & 72.2&	83.5&	12.8&	67.8&	73.3&	64.0&	66.5&	84.2&	85.8&	46.5	&35.7&	84.1&	56.4&	30.0&	31.1	&59.6\\
&
BFTT3D\,$\dagger$ \cite{wang2024backpropagation} & 85.5&	81.7&	19.3&	68.1&	85.2&	71.0&	72.8&	87.2	&  89.5 &	60.3	& 31.4  &	85.3&	66.0&	45.9&44.1	&	66.2\\
\cmidrule{2-19}
& 
MEMO \cite{zhang2022memo} & 72.5&	83.4&	11.0	& 66.2	& 72.4 &	63.2&  65.5 &	83.6  &	85.4	& 47.0 &	35.2 &	80.4	& 52.3 &	33.0 &	26.6&  58.5   \\
&
TENT \cite{wang2020tent} & 72.4&	83.5&	\textbf{13.1}	&68.4	&\textbf{74.1}&	65.0&	68.3&	84.1&	86.3	&48.1&	37.6&	\textbf{84.5}	&57.8&	30.4&	32.9&   60.4   \\
& 
\ccol  Ours & \ccol \textbf{73.1} &	\ccol \textbf{84.0} &	\ccol 12.1 &	\ccol \textbf{69.0} &	\ccol 73.5 &	\ccol \textbf{66.1} &	\ccol \textbf{68.6} &	\ccol \textbf{84.5} &	\ccol \textbf{86.1} &	\ccol \textbf{49.2} &	\ccol \textbf{40.8} &	\ccol 84.0 &	\ccol \textbf{59.7} &	\ccol \textbf{34.8} &	\ccol \textbf{34.2} &   \ccol \textbf{61.3}        \\
\midrule
\midrule
\parbox[t]{4mm}{\multirow{6}{*}{\rotatebox[origin=c]{90}{DGCNN}}} & 
Source-Only & 68.1 &	74.9&	13.6&	55.1&	74.7 &	74.7 &	75.3 &	51.5 &	82.2 &	76.7 &	57.2 &	57.6 &	76.4 &	32.2 &	12.0&       58.8         \\
% LAME & DGCNN & 23.4 & 15.3 & 4.0 & 4.5 & 32.2 & 6.6 & 8.8 & 40.1 & 65.2 & 3.5 & 2.5 & 31.48 & 7.5 & 4.0 & 4.0 & 16.8 \\
&
SHOT \cite{liang2020we} & 75.3&	78.2	&29.1&	66.6&	76.0&	69.9&	68.9&	50.4&	68.8&	59.4&	44.7&	42.1	&47.1&	12.5	&7.4	&53.1&                   \\
&
DUA \cite{mirza2022norm} & 81.6	&83.1	&39.4&	74.0&	83.4	&81.2&	82.1	&71.4&	85.5&	81.2&	72.3&	74.3&	80.5&	39.0 &	25.5&	70.3      \\
&
PL \cite{lee2013pseudo} & 79.6&	81.9&	35.4&	73.6&	79.1&	80.0&	80.8&	78.9	&86.8&	81.1	&72.1&	80.0&	79.6&	38.2&	33.8&	70.7\\
&
BFTT3D\,$\dagger$ \cite{wang2024backpropagation} & 80.5&	80.0&	41.5&	77.7&	75.4&	78.0&	79.3&	76.0	& 83.5 &	81.2	&  68.5 &	78.4&	78.6&	43.7&	37.8&	70.7\\
\cmidrule{2-19}
& 
MEMO \cite{zhang2022memo} & 79.0&	80.7&	50.1	& 67.9	& 78.7 &	74.1&  74.6 &	75.7  &	77.8	& 66.0 &	57.9 &	62.6	& 56.3 &	30.7 &	20.5&  63.5   \\
&
TENT \cite{wang2020tent} & \textbf{80.4}	& \textbf{82.0} &	38.6	&\textbf{75.2}	&80.1	&\textbf{80.8}	&\textbf{81.1}	&78.7 &	86.0	&\textbf{81.4}&	74.3 &	80.8 &	\textbf{80.5} &	39.2	&34.9	&71.6    \\
& 
\ccol
Ours & \ccol 80.1 &	\ccol 81.4	&\ccol \textbf{56.8}	&\ccol 73.2 &	\ccol \textbf{83.1} &	\ccol 79.1 &	\ccol 80.6 	& \ccol\textbf{81.9}	& \ccol\textbf{86.6}	& \ccol 79.3	& \ccol \textbf{74.5} &	\ccol \textbf{81.6} &	\ccol 79.0 &	\ccol \textbf{51.9} &	\ccol \textbf{43.2} &	\ccol \textbf{74.2}\\
\bottomrule
\end{tabular}}
\caption{Top-1 Classification Accuracy (\%) for all distribution shifts in the ModelNet-40C dataset. $*$ and $\dagger$ are explained in Section~\ref{main_results}.}
\label{tab:modelnet40c}
\end{table*}

\begin{table*}[t!]
\setlength\tabcolsep{4pt}
\centering
\resizebox{0.9\textwidth}{!}{
\small
\begin{tabular}{l|l|ccccccccccccccc|cc}
\toprule
& Method & \rotatebox{60}{uni} & \rotatebox{60}{gauss} & \rotatebox{60}{backg} & \rotatebox{60}{impul} & \rotatebox{60}{upsam} & \rotatebox{60}{rbf} & \rotatebox{60}{rbf-inv} & \rotatebox{60}{den-dec} & \rotatebox{60}{dens-inc} & \rotatebox{60}{shear} & \rotatebox{60}{rot} & \rotatebox{60}{cut} & \rotatebox{60}{distort} & \rotatebox{60}{oclsion} & \rotatebox{60}{lidar} & \rotatebox{0}{Mean} & \\
\midrule
\parbox[t]{4mm}{\multirow{9}{*}{\rotatebox[origin=c]{90}{Point-MAE}}} & 
Source-Only & 77.4&	71.8&	8.6	&54.4&	77.9	&75.5&	76.0	&85.3	&76.5&	80.5&	57.1	&85.1&	76.0	&11.0	&7.1&         61.3    \\
&
DUA* \cite{mirza2022norm} & 76.1 & 70.1 &  14.3 & 60.9 & 76.2 & 71.6 &  72.9 & 80.0 & 83.8 & 77.1 & 57.5 & 75.0 & 72.1 &  11.9 & 12.1 &   60.8 \\
& 
TTT-Rot* \cite{sun2020test} & 74.6 &72.4& 23.1 &59.9 &74.9& 73.8 &75.0& 81.4 &82.0& 69.2& 49.1& 79.9& 72.7 &14.0 &12.0&    60.9        \\ 
& 
T3A* \cite{iwasawa2021test} &  70.0 &60.5& 6.5& 40.7&  67.8& 67.2& 68.5 &79.5& 79.9& 72.7& 42.9& 79.1& 66.8& 7.7& 5.6&     54.4   \\
&
SHOT \cite{liang2020we} & 78.6	&74.2	&10.4 &	62.3	&73.9	&68.3	&64.9 &	68.1	&53.0 &	52.5	&31.4 &	54.2 &	41.2 &	2.3 &	1.7 &       49.1           \\
&
MATE* \cite{mirza2023mate} & 77.8& 74.7& 4.3& 66.2& 78.6& 76.3& 75.3 &86.1& 86.6& 79.2& 56.1& 84.1& 76.1& 12.3& 13.1& 63.1 \\
&
PL \cite{lee2013pseudo} & 80.8&	78.4&	16.4&	71.4&	81.2&	79.5&	79.8&	85.3&	83.2&	81.6&	69.4&	84.9&	79.5&	10.7&	10.3&	66.2    \\
\cmidrule{2-19}
& MEMO \cite{zhang2022memo} & 81.2&	74.5&	17.5	& 60.3 &	63.8 &	58.9 & 54.0 &	50.2 &	40.4	& 33.8 & 25.6 	& 23.4 &	18.4 &	16.6 &	16.4 &    42.3                 \\
& TENT \cite{wang2020tent} & 81.1&	78.7&	14.3	&\textbf{71.4} &	81.6 &	79.4 &	79.7 &	\textbf{85.9} &	82.6	&\textbf{81.8} &	70.0 	&\textbf{85.0} &	\textbf{79.6} &	9.9 &	10.4 &    66.1                 \\
& \ccol Ours & \ccol \textbf{82.6}	& \ccol \textbf{81.0}	& \ccol \textbf{21.2} &	\ccol 71.2	&\ccol \textbf{82.5} &	\ccol \textbf{79.8} &	\ccol \textbf{80.1}&	\ccol 85.2&	\ccol \textbf{84.8} &	\ccol 81.1&	\ccol \textbf{70.4}&	\ccol 83.9&	\ccol 78.9	&\ccol \textbf{11.2}&	\ccol \textbf{11.5}	&\ccol \textbf{67.0}     \\
\midrule
\midrule
\parbox[t]{4mm}{\multirow{6}{*}{\rotatebox[origin=c]{90}{PointNet}}} &
Source-Only & 59.9&	76.1&	9.3&	52.8&	60.3	&55.3 &	55.0 &	83.1	&82.6 &	42.9 &	26.3 &	83.0 &	47.8 &	6.3 &	5.5&	49.8     \\
%LAME & PointNet & 23.4 & 15.3 & 4.0 & 4.5 & 32.2 & 6.6 & 8.8 & 40.1 & 65.2 & 3.5 & 2.5 & 31.48 & 7.5 & 4.0 & 4.0 & 16.8 \\
&
SHOT \cite{liang2020we} & 65.5 &	77.7	&6.4 &	39.2 &	37.9 &	27.6 	&25.6 &	51.2 	&39.0 	&9.0 &	7.1 &	39.3 &	11.8&	1.8 &	2.1 &	29.4    \\
&
DUA \cite{mirza2022norm} & 66.7	&78.6	&12.5&	57.1	&66.4	&59.6&	60.9&	83.1	&82.4&	46.1&	34.1&	83.0& 	52.7	&8.2	&9.6&	53.4     \\
&
PL \cite{lee2013pseudo} & 67.0&	78.7&	11.8&	57.4&	67.8&	61.5&	62.2&	83.2&	82.8	&48.5&	37.6&	83.0&	54.9&	8.5&	9.0&	54.3    \\
\cmidrule{2-19}
&
MEMO \cite{zhang2022memo} & 65.9&	77.0&	\textbf{13.9}	& 42.9	& 36.4 &	27.9&  21.9 &	17.8  &	16.7	& 17.3 &	16.6 &	16.4	& 16.5 &	\textbf{17.1} &	\textbf{16.0}&  28.0   \\
&
TENT \cite{wang2020tent} & 67.2 &	79.1 &	12.8	&59.2 &	67.8	&62.9 &	63.6 &	\textbf{83.4} &	\textbf{82.8} &	51.6	&40.9	&\textbf{83.0} &	58.2 &	9.6	&9.5 &	55.5      \\
& 
\ccol Ours & \ccol \textbf{68.3}&	\ccol \textbf{79.3} &	\ccol 12.4 &	\ccol \textbf{61.2} &	\ccol \textbf{70.7}	&\ccol \textbf{65.8} &	\ccol \textbf{66.8} &	\ccol 82.2 &	\ccol 82.3	&\ccol \textbf{56.6}	&\ccol \textbf{46.0} 	&\ccol 80.9	&\ccol \textbf{60.4} &	\ccol 10.0 &	\ccol 10.1 &	\ccol \textbf{56.9}      \\
\midrule
\midrule
\parbox[t]{4mm}{\multirow{6}{*}{\rotatebox[origin=c]{90}{DGCNN}}} & 
Source-Only & 74.7 &	74.1 	&30.5	&55.9	&75.0 &	77.3	&78.1&	85.1 &	81.6	&79.2 &	66.2	&84.7 &	77.4	&8.0 &	6.5 &	63.6      \\
% LAME & DGCNN & 23.4 & 15.3 & 4.0 & 4.5 & 32.2 & 6.6 & 8.8 & 40.1 & 65.2 & 3.5 & 2.5 & 31.48 & 7.5 & 4.0 & 4.0 & 16.8 \\
& 
SHOT \cite{liang2020we} & 73.1 &	68.2&	14.6 &	52.2 &	48.9 &	43.0 &	36.3	&43.8	&27.8 &	22.4 &	15.4&	25.0 &	14.6 &	1.9 &	1.8	&32.6     \\
& 
DUA \cite{mirza2022norm} & 78.1 &	77.4	&24.5&	72.8	&78.1	&79.9&	80.4&	85.3&	82.5	&81	&73.3	&84.6&	79.1	&9.3	&10.5	&66.4 \\
&
PL \cite{lee2013pseudo} & 78.3&	78.2&	24.1&	73.4	&79.3	&80.4	&81.1&	85.5&	82.7&	81.5&	74.5&	84.7&	80.0&	10.1&	11.6&	67.0   \\
\cmidrule{2-19}
&
MEMO \cite{zhang2022memo} & 77.5&	69.0 &	34.0	& 40.0	& 27.3 & 23.9  &  20.6 &	18.1  &	17.2	& 17.0 &	16.9 &	16.7	& 16.7 &	\textbf{18.1} &	\textbf{16.8} &  28.7   \\
&
TENT \cite{wang2020tent} & 78.7 	&78.7 &	27.7 	&\textbf{73.5} 	&\textbf{78.6} &	\textbf{79.8} &	\textbf{80.2} &	\textbf{84.4} &	\textbf{81.8} &	\textbf{79.9} &	\textbf{74.6}	&\textbf{83.7}	&\textbf{79.2} 	& 10.6	& 14.2 &	\textbf{67.0}     \\
&
\ccol Ours & \ccol \textbf{80.0} &	\ccol \textbf{80.0} &	\ccol \textbf{59.0} 	& \ccol 69.5	&\ccol 77.0 	&\ccol 74.9 &	\ccol 74.7	&\ccol 80.0 &	\ccol 78.9 	& \ccol 74.9 &	\ccol 70.5 &	\ccol 77.9 &	\ccol 72.4 &	\ccol 9.1 &	\ccol 12.0 &	\ccol 66.0    \\
\bottomrule
\end{tabular}}
\caption{Top-1 Classification Accuracy (\%) for all distribution shifts in the ShapeNet-C dataset. $*$ is explained in Section~\ref{main_results}.}
\label{tab:ShapeNet}
\end{table*}

\begin{table*}[t!]
\setlength\tabcolsep{4pt}
\centering
\resizebox{0.9\textwidth}{!}{
\small
\begin{tabular}{l|l|ccccccccccccccc|cc}
\toprule
&
Method & \rotatebox{60}{uni} & \rotatebox{60}{gauss} & \rotatebox{60}{backg} & \rotatebox{60}{impul} & \rotatebox{60}{upsam} & \rotatebox{60}{rbf} & \rotatebox{60}{rbf-inv} & \rotatebox{60}{den-dec} & \rotatebox{60}{dens-inc} & \rotatebox{60}{shear} & \rotatebox{60}{rot} & \rotatebox{60}{cut} & \rotatebox{60}{distort} & \rotatebox{60}{oclsion} & \rotatebox{60}{lidar} & \rotatebox{0}{Mean} & \\
\midrule
\parbox[t]{4mm}{\multirow{9}{*}{\rotatebox[origin=c]{90}{Point-MAE}}} & 
Source-Only & 20.8 &	32.5 &	16.7 	&16.0 &	22.4	&32.2	&35.3 &	68.8 &	64.9 &	35.8	&28.6	&69.7 &	33.9 &	9.3 &	9.1 &	33.1       \\
&
DUA* \cite{mirza2022norm} & - & - &  - & - & - & - &  - & - & - & - & - & - & - &  - & - &   46.0 \\
& 
TTT-Rot* \cite{sun2020test} & - & - &  - & - & - & - &  - & - & - & - & - & - & - &  - & - &    46.1        \\ 
&
T3A* \cite{iwasawa2021test} & - & - &  - & - & - & - &  - & - & - & - & - & - & - &  - & - &     40.3   \\
&
SHOT \cite{liang2020we} & 38.7	&53.7 &	17.0 &	31.4 	&37.9	&47.3	&46.5	&71.7	&70.7 &	50.0 &	45.5 &	70.3 &	49.0	&9.4 &	8.8 &  43.2\\
&
PL \cite{lee2013pseudo} & 38.1	&56.0 &	17.4&	31.4&	39.6&	51.0&	53.7&	74.6&	74.4&	54.7&	49.0&	75.4&	56.0&	9.0& 	7.4&	45.8   \\
&
MATE* \cite{mirza2023mate} & - & -& -& -& -& -& -& -& -& -& -& -& -& -& - & 47.0\\
\cmidrule{2-19}
&
MEMO \cite{zhang2022memo} & 40.4	& 55.5	& 18.6	& \textbf{34.0} 	& \textbf{41.4}	& 51.8	& 53.3	& 73.6 &	74.6	& 54.5 &	46.7	& 73.0	& 55.7 &	8.8 &	\textbf{9.2} &	46.1    \\
&
TENT \cite{wang2020tent} & 38.1	&55.8	&16.6	&32.4	&39.6	&51.0	&\textbf{54.7}	&\textbf{74.2} &	\textbf{75.0}	&54.5 &	\textbf{50.0}	&\textbf{74.2}	&\textbf{56.8} &	\textbf{8.8} &	7.6 &	45.9    \\
&
\ccol Ours &  \ccol \textbf{41.0}&	\ccol \textbf{58.8}	&\ccol \textbf{18.7} &	\ccol 33.6	&\ccol 39.6 &	\ccol \textbf{51.9}	&\ccol 52.5 &	\ccol 72.1 &	\ccol 74.2 &	\ccol \textbf{55.7} &	\ccol 48.2	&\ccol 74.0 &	\ccol 54.3&	\ccol 8.2 &	\ccol 9.0 &	\ccol \textbf{46.1}    \\
\midrule
\midrule
\parbox[t]{4mm}{\multirow{6}{*}{\rotatebox[origin=c]{90}{PointNet}}} &
Source-Only & 20.6 &	36.1 &	10.3 &	18.4 &	20.6 	&25.6 &	28.4	&63.2 &	64.5 &	27.9 &	23.1 &	62.8 &	27.0 &	6.9&	10.0 &	29.7   \\
%LAME & PointNet & 23.4 & 15.3 & 4.0 & 4.5 & 32.2 & 6.6 & 8.8 & 40.1 & 65.2 & 3.5 & 2.5 & 31.48 & 7.5 & 4.0 & 4.0 & 16.8 \\
&
SHOT \cite{liang2020we} & 42.2 &	62.3 &	15.0 	&33.8	&41.6	&35.9 &	35.3 &	66.0 &	66.4 &	36.3 &	32.8 &	65.4 	&40.4 &	9.0 &	8.4&	39.4      \\
& 
DUA \cite{mirza2022norm} & 28.1&	43.3	&13.9&	21.7&	31.2	&29.1&	34.8&	62.9&	65.6&	30.1&	26.0& 	63.3&	31.2	&7.6&	9.2	&33.2        \\
&
PL \cite{lee2013pseudo} & 40.6&	61.5&	16.2&	36.5	&40.4&	35.5&	36.1&	64.8	&65.4&	36.3&	35.1&	66.8&	39.6&	8.0&	9.4	&39.5    \\
&
BFTT3D\,$\dagger$ \cite{wang2024backpropagation} & 42.9&	60.1&	30.0&	34.8&	44.1 &	45.4&	47.0&	78.7	& 78.5 & 41.3 &	31.2	&  75.9 &	44.1&	15.0&	14.5 &	45.5\\
\cmidrule{2-19}
&
MEMO \cite{zhang2022memo} & 41.6	& \textbf{62.5}	& \textbf{17.6}	& \textbf{36.1} 	& \textbf{41.2}	& \textbf{37.5}	& \textbf{37.7}	& \textbf{67.6} &	65.4	& 39.1 &	31.8	& 65.0	& 39.8 &	10.4 &	9.4 &	40.2    \\
&
TENT \cite{wang2020tent} & 40.2	&61.5 	&16.2 &	35.7 &	40.8 &	35.7	&37.1 &	65.4	&66.4 &	35.9	&\textbf{35.1} 	&\textbf{67.0} 	&39.6 	&8.8 &	\textbf{10.1}	&39.7   \\
&
\ccol Ours & \ccol \textbf{42.0} &	\ccol 61.9 &	\ccol 15.8 &	\ccol 35.9 &	\ccol 41.0 &	\ccol 37.1 &	\ccol 37.3 &	\ccol 67.4 &	\ccol \textbf{68.2} &	\ccol \textbf{40.6}	&\ccol 35.0 &	\ccol 66.8 &	\ccol \textbf{40.2}	&\ccol \textbf{10.5} 	&\ccol 9.2 &	\ccol \textbf{40.6 }     \\
\midrule
\midrule
\parbox[t]{4mm}{\multirow{6}{*}{\rotatebox[origin=c]{90}{DGCNN}}} &
Source-Only & 26.7 &	39.9 	&27.0 &	22.5 &	29.6 &	41.5 &	41.3 &	70.6 &	60.2 	&43.4 &	34.4 &	70.2 	&43.0 &	8.7 &	9.3 &	37.9      \\
% LAME & DGCNN & 23.4 & 15.3 & 4.0 & 4.5 & 32.2 & 6.6 & 8.8 & 40.1 & 65.2 & 3.5 & 2.5 & 31.48 & 7.5 & 4.0 & 4.0 & 16.8 \\
&
SHOT \cite{liang2020we} & 36.8 &	48.9 &	28.1 &	45.8 	&33.8 	&50.7 &	49.6 &	70.8 &	65.1 &	53.3 &	47.4 &	67.5 &	51.7 	&8.7 	&8.8 &	44.5      \\
&
DUA \cite{mirza2022norm} &  38.4&	51.9&	30.2&	48.4&	39.2&	54.0& 	55.2&	73.6&	70.7&	57.8&	50.3&	74.3	&56.9&	9.2&	10.8&	48.1 \\
&
PL \cite{lee2013pseudo} & 38.4	& 53.6&	27.6&	52.1&	39.1&	55.0&	58.0&	73.6&	68.2&	57.5&	53.3&	71.3&	57.3&	9.2&	8.1&	48.1 \\
&
BFTT3D\,$\dagger$ \cite{wang2024backpropagation} & 42.7&	58.2&	47.7&	56.6&	46.6&	63.2&	65.9&	77.5	& 82.3 &	67.0	&  61.5 &	77.8&	65.4&	14.5&	14.6&	56.1\\
\cmidrule{2-19}
& 
MEMO \cite{zhang2022memo} & 37.7	& 52.6	& \textbf{28.0}	& \textbf{52.4} 	& 39.2	& 51.2	& 54.2	& 67.2 &	62.0	& 48.4 &	47.9	& 59.5	& 50.3 &	\textbf{11.3} &	\textbf{13.0} &	45.0  \\
&
TENT \cite{wang2020tent} & 39.1 &	\textbf{55.2} &	25.3 &	50.5 &	39.7 	&54.0 	&59.2 &	\textbf{73.4} &	68.4 &	58.8 &	54.2 &	71.9 &	56.2 	&10.1 	&8.3 	&48.3      \\
&
\ccol Ours & \ccol \textbf{39.2} &	\ccol 54.2	&\ccol 27.8	&\ccol 51.6 &	\ccol \textbf{43.0} 	&\ccol \textbf{58.3} &	\ccol \textbf{59.9} &	\ccol 70.5 &	\ccol \textbf{71.2} &	\ccol \textbf{61.4} 	&\ccol \textbf{55.4} 	&\ccol \textbf{74.1} 	&\ccol \textbf{60.6}	&\ccol 10.2 &	\ccol 10.6 &	\ccol \textbf{49.9}     \\
\bottomrule
\end{tabular}}
\caption{Top-1 Classification Accuracy (\%) for all distribution shifts in the ScanObjectNN-C dataset. $*$ and $\dagger$ are explained in Section~\ref{main_results}.}
\label{tab:scann}
\end{table*}

\subsection{Main Results}
\label{main_results}
In all result tables, \textit{source only} refers to testing the pretrained model directly on the corrupted dataset without any adaptation. While BFTT3D \cite{wang2024backpropagation} is included in the tables, its results are not directly comparable to ours because since, as mentioned in the Introduction, this method relies on source data during \gls{tta}. Except for the ones marked with $*$, all results are reproduced. Moreover, results marked with $\dagger$ indicate dependence on the source data during \gls{tta}.

\mypar{ModelNet-40C.} 
We evaluated the effectiveness of our method on the ModelNet40-C dataset using three different backbones, Point-MAE, PointNet, and DGCNN, under various corruptions. As reported in Table~\ref{tab:modelnet40c}, our method consistently outperforms prior approaches across different backbones and corruptions. For the Point-MAE backbone, our method improves performance across all corruption types, with the average accuracy increasing from 72.1\% (TENT) and 69.4\% (MEMO) to 75.0\%. This reflects the robustness of our approach in handling distribution shifts by leveraging the proposed sampling variation and weight averaging technique, as discussed in Section~\ref{sec:method}. 
In the PointNet backbone, our method achieves an overall improvement in mean accuracy from 60.4\% (TENT) 58.5 (MEMO) to 61.3\%. This improvement is driven by the better handling of difficult corruptions such as \textit{rotation}, \textit{distortion}, and \textit{occlusion}, where our method consistently performs better than TENT. For the DGCNN backbone, our method performs exceptionally well on specific corruptions such as \textit{occlusion} and \textit{lidar}, achieving improvements of 12.7\% and 8.3\%, respectively, over TENT, and 21.2\% and 22.7\%, respectively, over MEMO. This leads to a higher overall average, increasing from 71.6\% (TENT) and 63.5\% (MEMO) to 74.2\%. These results highlight the effectiveness of our method in improving robustness under various corruption scenarios, particularly when adapting to real-world data shifts during test-time adaptation.

\mypar{ShapeNet-C.} 
In Table~\ref{tab:ShapeNet}, we evaluate our method on the ShapeNet-C dataset across the same three backbones. For the Point-MAE backbone, our method outperforms TENT and MEMO, achieving a mean accuracy of 67.0\% compared to TENT's 66.1\% and MEMO's 42.3\%. Particularly large gains in performance are obtained by our method for the \textit{background} and \textit{dens-inc} corruptions.
% In the PointNet backbone, we observe a similar trend, with our method yielding a mean accuracy of 56.9\%, outperforming TENT's 55.5\%. This improvement is particularly notable in difficult corruption scenarios such as \textit{background changes} (12.4\%) and \textit{upsampling} (65.8\%), demonstrating the effectiveness of our approach in enhancing generalization across distribution shifts.
We observe similar improvements with the PointNet backbone, as our method achieves a mean accuracy of 56.9\%, surpassing TENT's score of 55.5\% and MEMO's score of 28.0\%. Our model shows significant performance gains in difficult corruptions like \textit{upsampling}, \textit{rbf}, \textit{rotation}, \textit{shear}, and \textit{rbf-inv}. These results demonstrate the effectiveness of our method in improving robustness across various corruption types, ensuring the model generalizes better to corrupted data during test-time adaptation. For the DGCNN backbone, our method outperforms MEMO, although its performance is somewhat comparable to the TENT method.

% For the \textbf{DGCNN} backbone, significant improvements are achieved in \textit{occlusion} (12.0\%) and \textit{lidar} (9.1\%), resulting in an overall mean accuracy of \textbf{66.0\%}, demonstrating the robustness of our method in diverse corruptions.

\mypar{ScanObjectNN-C.} 
As a final evaluation, we assessed our method on the ScanObjectNN-C dataset, using three backbones. As presented in Table \ref{tab:scann}, our method achieved notable improvements in performance across multiple corruptions and backbones. 
% In the Point-MAE backbone, our approach improves mean accuracy to 46.1\%, surpassing TENT's 45.9\%, with notable gains in corruptions like \textit{Gaussian noise} (58.8\%) and \textit{uniform noise} (41.0\%).  
For the Point-MAE backbone, our approach improves the mean accuracy to 46.1\%, surpassing TENT's 45.9\%, with notable gains in corruptions such as \textit{Gaussian noise} (58.8\%) and \textit{uniform noise} (41.0\%). On this dataset with this backbone, our performance is equal to MEMO.
Likewise, our method improved the mean accuracy to 40.6\% for the PointNet backbone, outperforming TENT's 39.7\%. This improvement is particularly evident in challenging corruptions such as \textit{occlusion} (10.5\%) and \textit{shear} (40.6\%). On this dataset and with this backbone, MEMO performs comparably to our method.
In the DGCNN backbone, our approach yielded a mean accuracy of 49.9\%, outperforming TENT's 48.3\% and MEMO's 45.0\%. Our method showed strong improvements like in \textit{upsample} (43.0\%), \textit{rbf} (58.3\%), and \textit{shear} (61.4\%), further demonstrating its robustness in handling distribution shifts across the dataset.

\subsection{Ablation Study}
In this section, we conduct a comprehensive ablation study on the ModelNet40-C dataset using the Point-MAE backbone to examine the impact of various factors on our model's performance in \gls{tta}.
% In this section, we perform a comprehensive ablation study on ModelNet40-C dataset to investigate how various factors affect our model's performance in \gls{tta}. 
Specifically, we evaluate four aspects: \emph{Sampling Variation}, \emph{Number of Iterations}, \emph{Batch Size}, and \emph{Types of Augmentation}. For consistency, all experiments were conducted with a learning rate of 0.001.

\begin{figure}
  \centering
  \begin{minipage}[t]{0.48\textwidth}
    \includegraphics[width=\linewidth]{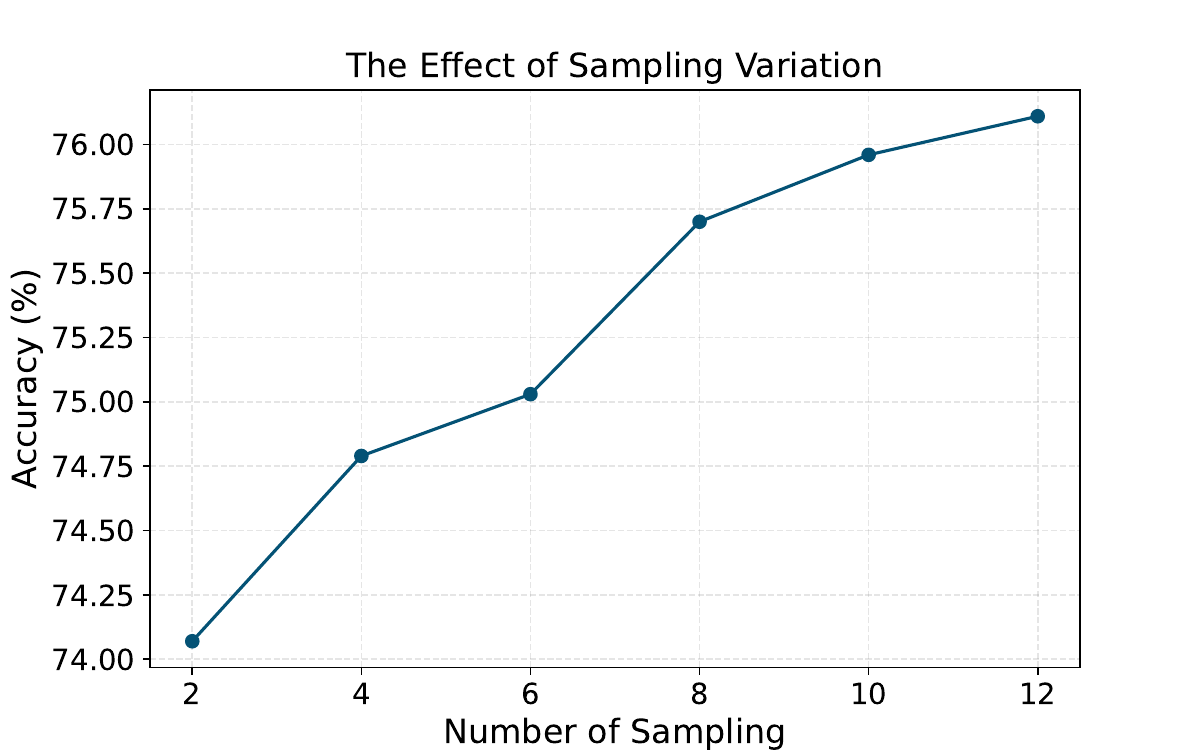} % Set width to \linewidth or less to ensure it fits within the column and is centered
    \caption{Impact of Different $N^V$ on Model Accuracy}
    %Point-MAE, Point-MAE+\gls*{geomask3d}\textsuperscript{*} and Point-MAE+\gls*{geomask3d}
    \label{fig:sampling_variation}
  \end{minipage}\hfill
  \vspace{8pt}
  \begin{minipage}[t]{0.48\textwidth} % Adjust width to your preference
    \includegraphics[width=\linewidth]%{graphics/POINTMAE_NEW_BMVC2024_size_18__OBJBG_70_3_models_fontsize.pdf}
    {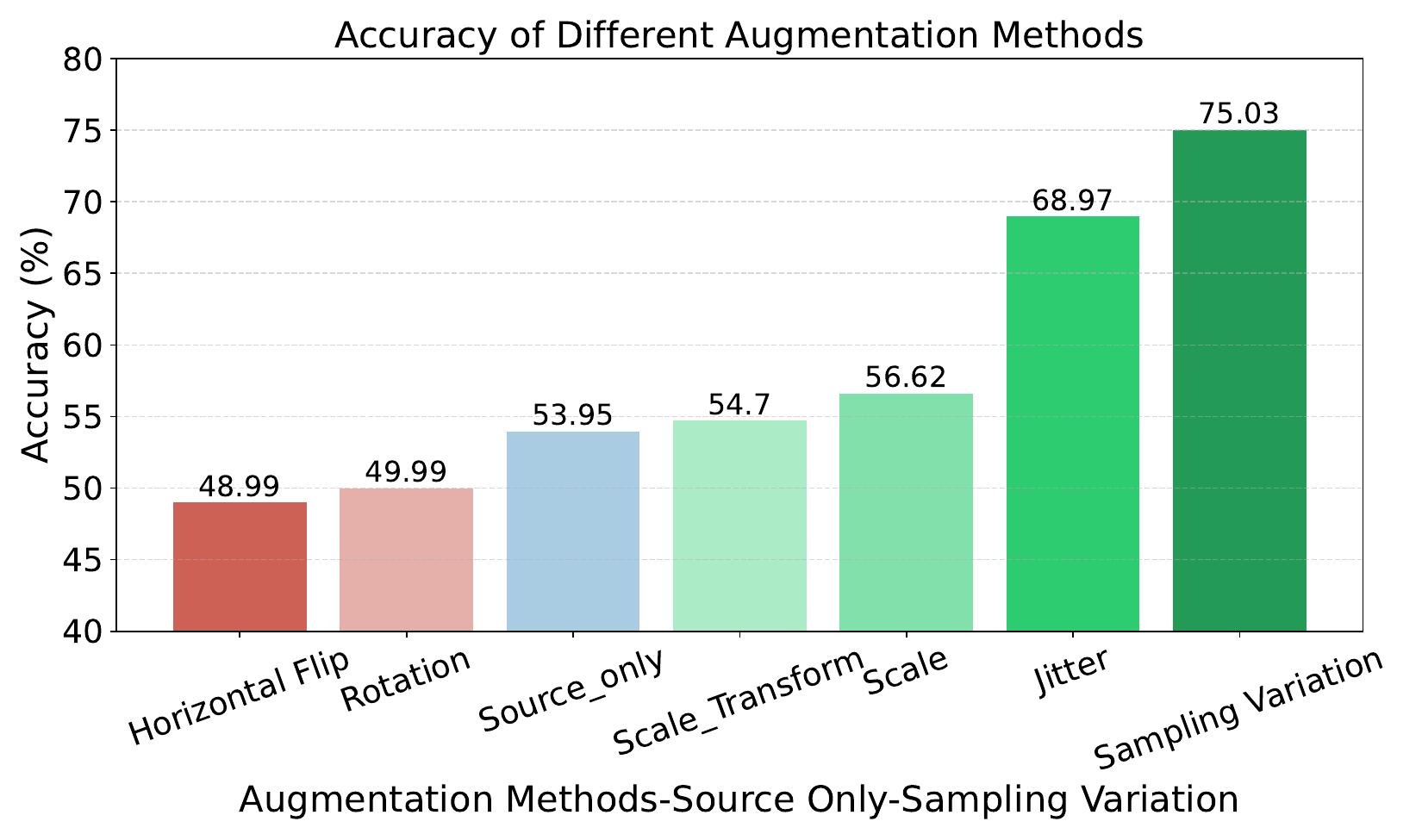}
    \caption{Comparison between Sampling Variation and Different Augmentations}
    \label{fig:augmentation}
  \end{minipage}
\end{figure}

\mypar{Sampling Variation.} 
We first explore the effect of increasing the number of sampling variations $N^V$. Figure~\ref{fig:sampling_variation} shows the steady rise in model accuracy as we increase the number of sampling variations. This validates our idea of combining weight averaging and sampling variation to enhance the model's robustness. As 
$N^V$ increases from 2 to 12, the accuracy improves approximately from 74.0\% to 76.0\%. This demonstrates that leveraging more diverse sampling during TTA enables better adaptation to distribution shifts. Based on these results, we selected $N^V\!=\!6$ for all subsequent experiments to balance computational efficiency and performance improvement. For this experiment, the batch size and iteration are set to 128 and 1, respectively.

\begin{figure}
  \centering
  \begin{minipage}[!t]{0.48\textwidth}
    \includegraphics[width=\linewidth]{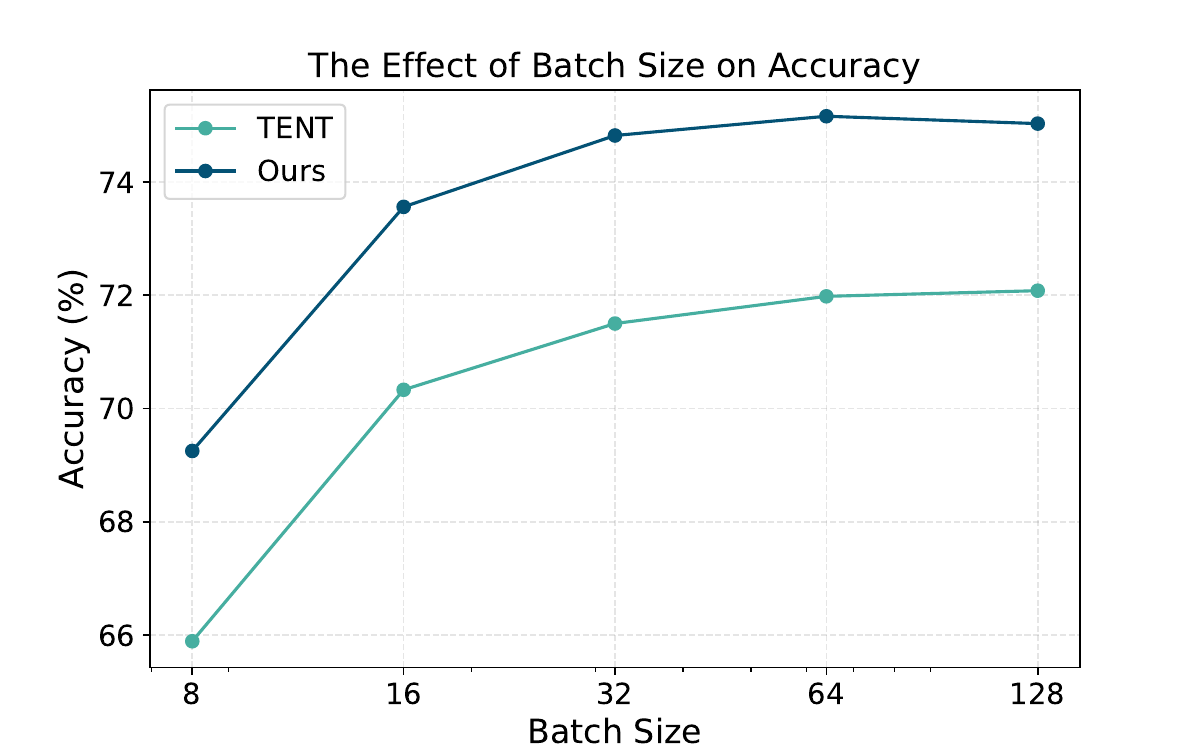} % Set width to \linewidth or less to ensure it fits within the column and is centered
    \caption{Impact of Batch Size on Accuracy for Two Methods}
    %Point-MAE, Point-MAE+\gls*{geomask3d}\textsuperscript{*} and Point-MAE+\gls*{geomask3d}
    \label{fig:batch_size}
  \end{minipage}\hfill
  \begin{minipage}[!t]{0.48\textwidth} % Adjust width to your preference
    \includegraphics[width=\linewidth]%{graphics/POINTMAE_NEW_BMVC2024_size_18__OBJBG_70_3_models_fontsize.pdf}
    {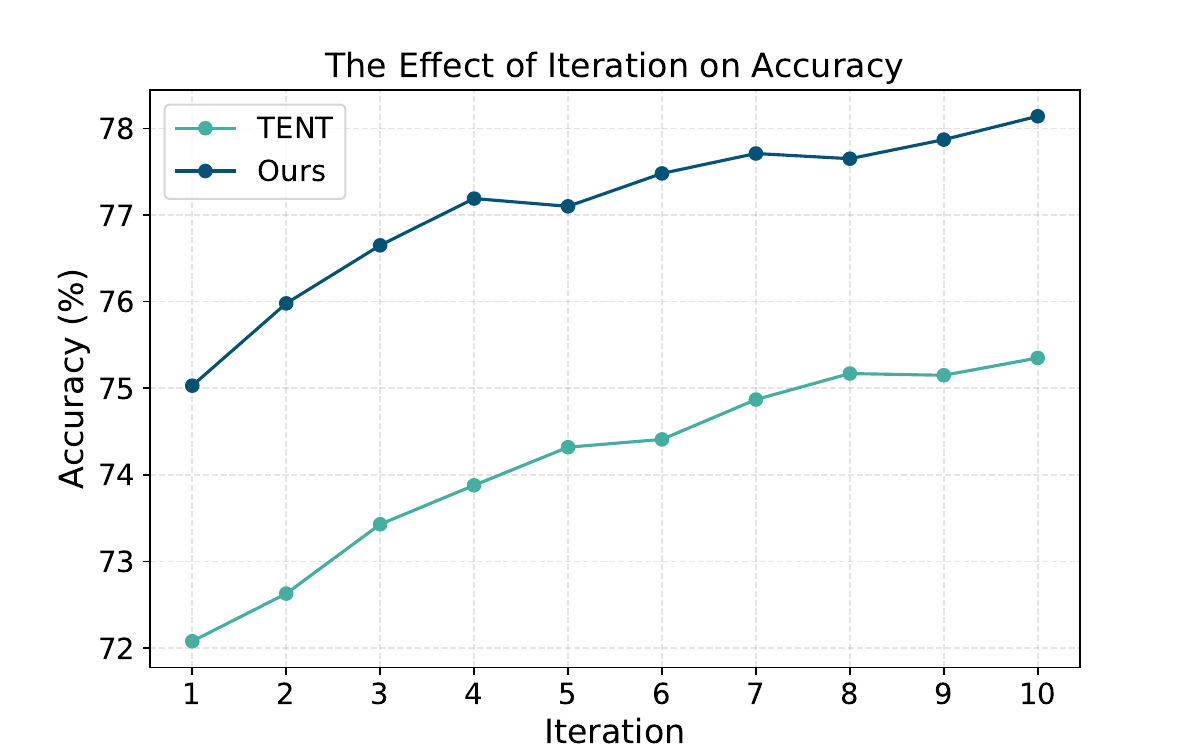}
    \caption{Impact of Iteration on Accuracy for Two Methods}
    \label{fig:iteration_effect}
  \end{minipage}
\end{figure}

%\vspace{5pt}
\mypar{Number of Iterations.}
As shown in Figure \ref{fig:iteration_effect}, the accuracy improves as the number of iterations increases. Our method and TENT show performance gains with more iterations, but our method consistently outperforms TENT at every stage of the iteration process. Starting with a gap at the first iteration, our approach maintains a steady improvement, surpassing TENT at each level.  Based on these results, we opted to use just one iteration in subsequent experiments to prioritize faster model adaptation while still achieving a competitive accuracy boost. For this experiment, the batch size is set to 128.

%\vspace{5pt}
\mypar{Batch Size.}
As illustrated in Figure~\ref{fig:batch_size}, our method consistently outperforms TENT across all batch sizes. Notably, the improvement remains steady even at lower batch sizes, such as 8 and 16, where our method achieves approximately 3\% higher accuracy. This demonstrates that our approach is robust across different batch settings, making it effective even in scenarios with limited data. 
%The consistent performance at both small and large batch sizes showcases the flexibility of our method in different experimental settings. 
Based on these results, we selected a batch size of 128. For this experiment, the number of iterations is set to 1.

\vspace{5pt}
\mypar{Types of Augmentation.}
It can be argued that the effect of sampling variation is akin to applying various augmentations to the data samples. Hence, we have assessed the integration of different augmentations into our weight averaging process.
As shown in Figure~\ref{fig:augmentation}, we applied augmentations such as \textit{Horizontal Flip}, \textit{Rotation}, \textit{Scale}, \textit{Scale Transform}, \textit{Jitter}, and \textit{Sampling Variation} to investigate their impact. 

Suprisingly, certain augmentations like \textit{Horizontal Flip} and \textit{Rotation} worsened the model's performance, reducing accuracy even below the \textit{source-only} model.
However, augmentations like \textit{Jitter} and especially \textit{Sampling Variation} resulted in improvements, with \textit{Sampling Variation} outperforming all other strategies. This highlights that our approach, which leverages sampling variation combined with weight averaging, is highly effective for boosting performance compared to data augmentations. For this experiment, the batch size and iteration are set to 128 and 1, respectively.

% \vspace{5pt}
% \mypar{Parallel vs Sequential}
% In the context of the parallel and sequential methods illustrated in Figure~\ref{fig:parallel_sequential}, we can describe them as follows:

% - \textbf{Parallel Mode}: After adapting the model using each sampling variation \(\mathcal{P}_{\text{patch}}^i\), the model is reset to its initial state. The adapted weights from each sampling \(\theta_i\) are stored, and the final model weights \(\theta_{\text{avg}}\) are obtained by averaging all the adapted weights (Equation~\ref{eq:wa}).

% - \textbf{Sequential Mode}: In this approach, the model is not reset after each adaptation. Instead, the model continues adapting progressively, using the weights after each adaptation as the starting point for the next sampling variation \(\mathcal{P}_{\text{patch}}^i\), thereby adapting iteratively.

% As seen in Figure~\ref{fig:parallel_sequential}, the sequential mode offers similar performance as the number of sampling variations increases. However, since we aim for faster model adaptation, we select the parallel mode, as the model can be adapted in parallel with each sampling variation.

\section{Conclusion}
In this paper, we introduced a novel \gls{tta} framework combining weight averaging with sampling variation to enhance model robustness against distribution shifts in 3D point cloud data. Evaluated on multiple backbones and datasets, our method outperforms existing approaches such as TENT, particularly under challenging corruptions. Through ablation studies, we demonstrated the effectiveness of our approach across different batch sizes, iterations, and sampling variations. Our method offers a robust, efficient solution for improving generalization in 3D point cloud classification.

% \bibliographystyle{unsrt}
% \bibliography{main.bib} 

%%%%%%%%%%%%%%%%%%%%%%%%%

% \bibliographystyle{unsrt}
% \bibliography{main.bib} 

%%%%%%%%%%%%%%%%%%%%%%%%%%%%%%%%%%%%%%%%%%%%%%%%%%%%%%%%%%%%
\newpage

\section*{\centering Test-Time Adaptation in Point Clouds: Leveraging Sampling Variation with Weight Averaging - Supplementary Material \\ Supplementary Material}
\vspace{1cm}

\appendix

\section{Implementation}
We used PyTorch to implement the core functionalities of our approach. The codebase is structured into two main parts: \emph{Pretrain} and \emph{adaptation}.

\mypar{Pretrain.}
We begin by focusing on the initial pretraining phase of the base models (Point-MAE, PointNet, DGCNN, and CurveNet). During this phase, we pretrain the backbones in a fully supervised manner, following the standard definition of Test-Time Adaptation (TTA). The pretraining is conducted on clean datasets such as ModelNet, ShapeNet, and ScanObjectNN. This phase ensures that the models are adequately prepared for the subsequent adaptation steps.

\mypar{Adaptation.}
After completing the pretraining phase, we transition to the adaptation stage. In this phase, we only update the Normalization Layers of the models using our method, which is built upon the TENT algorithm. By selectively adapting the normalization layers, we efficiently adjust the models to handle corrupted data without requiring full retraining. This targeted approach not only reduces computational costs but also enhances the model’s ability to generalize to different data distributions. The results of this adaptation phase are directly reflected in the experimental findings presented in this paper.

% \mypar{Accessibility and Resources.}
To ensure complete transparency and reproducibility of our results, we have made all relevant materials publicly available. This includes:
\begin{itemize}\setlength\itemsep{.1em}
\item The full source code for both \emph{Pretrain} and \emph{Adaptation} phases;
%\item Configuration files necessary for replicating our experimental setup (log files). 
\item All log files containing the detailed results of our experiments;
\item Pretrained the base models for all backbones.
\end{itemize}

\begin{table}[b!]
\centering
\caption{Hyperparameters}
\label{tab:hyperparameters}
\small
\renewcommand{\arraystretch}{1.0} % Adjust the space between the table rows
\setlength{\tabcolsep}{1.7pt}
\begin{tabular}{llc}
\toprule
Backbone & Config & Value \\ 
\midrule
All & Optimizer & AdamW \\
All & learning rate & 1e-3 \\
All & Weight decay & 0.0 \\
All & Momentum & $\beta = 0.9$ \\
All & Iteration & 1 \\
All & FPS & 512, 1024 \\
PointMAE-PointNet & Batch size  & 128 \\
DGCNN-CurveNet & Batch size  & 16, 64 \\

\bottomrule
\end{tabular}

\end{table}

\begin{table*}[t]
\setlength\tabcolsep{4pt}
\centering
\resizebox{0.9\textwidth}{!}{
\small
\begin{tabular}{l|l|ccccccccccccccc|c}
\toprule
&
Method & \rotatebox{60}{uni} & \rotatebox{60}{gauss} & \rotatebox{60}{backg} & \rotatebox{60}{impul} & \rotatebox{60}{upsam} & \rotatebox{60}{rbf} & \rotatebox{60}{rbf-inv} & \rotatebox{60}{den-dec} & \rotatebox{60}{dens-inc} & \rotatebox{60}{shear} & \rotatebox{60}{rot} & \rotatebox{60}{cut} & \rotatebox{60}{distort} & \rotatebox{60}{oclsion} & \rotatebox{60}{lidar} & \rotatebox{0}{Mean}\\
\midrule
\parbox[t]{4mm}{\multirow{6}{*}{\rotatebox[origin=c]{90}{Point-MAE}}} &
Source-Only & 66.6 & 59.1 & 7.2 & 31.8 & 74.6 & 67.7 & 69.8 & 59.3 & 75.1 & 74.4 & 38.0 & 53.7 & 70.0 & 38.6 & 23.4 & 53.9 \\
&
\\
&
Ours (BN) &  85.4&	84.7 &	29.9&	74.8&	87.1&	80.9&	82.3&	85.1&	88.4 &	82.4&	67.9&	83.9&	80.7&	55.7&	54.8&	74.9\\								
&
\\
&

\ccol Ours (BN \& LN) & \ccol 85.0	& \ccol 83.9	&\ccol 33.0	&\ccol 74.6	&\ccol 87.0&	\ccol 80.9&	\ccol 82.3&	\ccol 85.1	&\ccol 88.0	&\ccol 82.7	&\ccol 66.9	&\ccol 84.0&	\ccol 80.5&	\ccol 56.2&	\ccol 55.3& \ccol 75.0  \\
& 
\\
% LAME & DGCNN & 23.4 & 15.3 & 4.0 & 4.5 & 32.2 & 6.6 & 8.8 & 40.1 & 65.2 & 3.5 & 2.5 & 31.48 & 7.5 & 4.0 & 4.0 & 16.8 \\
\bottomrule
\end{tabular}}
\caption{Top-1 Classification Accuracy (\%) for all distribution shifts in the ModelNet-40C dataset.}
\label{tab:bn_ln_results}
\end{table*}

All these resources can be accessed through our \href{https://github.com/AliBahri94/SVWA_TTA.git}{\emph{code}}. This repository includes everything needed to understand our code, covering all aspects of the implementations and the reproduction of the results. Moreover, the specific hyperparameters used for all backbones are comprehensively outlined in Table~\ref{tab:hyperparameters}.

\section{Resource Overhead}
\mypar{Time.} Our method builds on the TENT algorithm but extends it by introducing multiple sampling variations \( \mathcal{P}_v \) during \gls{tta}. While there may be concerns about potential resource overhead, particularly regarding execution time, our method is designed to run in parallel for all \( \mathcal{P}_v \). This parallelization allows the model to adapt independently for each variation, significantly reducing time costs compared to a sequential approach. The comparison between parallel and sequential adaptations is detailed in the Supplementary Material Section \ref{sec:add-exp}. 
To quantify the computational cost, we evaluated our method on the PointNet backbone, comparing it directly with TENT. Using $N^V\!=\!6$, the average adaptation time for TENT is approximately \textbf{21 ms}, whereas our method required around \textbf{26 ms}. This marginal difference indicates that the parallelization ensures minimal resource overhead, making our approach highly efficient even with multiple sampling variations.

\mypar{Memory.}
Given that our method adapts only the learnable parameters of the normalization layers, keeping the other weights frozen and shared, it involves a limited number of parameters in the adaptation process.
For instance, in the PointNet backbone, there are approximately 3,500,000 parameters, and we adapt only around 12,000 parameters, which constitutes \textbf{0.3\%} of all parameters. Consequently, when using $N^V\!=\!6$, the memory resource overhead is approximately \textbf{1.8\%} of the whole backbone, which is negligible.

\section{Additional Experiments}\label{sec:add-exp}
% \mypar{Parallel vs Sequential}
% In the context of the parallel and sequential methods illustrated in Figure~\ref{fig:parallel_sequential}, we can describe them as follows:

% - \textbf{Parallel Mode}: After adapting the model using each variation \(\mathcal{P}_v\), the model is reset to its initial state. The adapted weights from each variation \(\theta_v\) are stored, and the final model weights \(\theta_{\text{avg}}\) are obtained by averaging all the adapted weights (Equation 5).

% - \textbf{Sequential Mode}: In this approach, the model is not reset after each adaptation. Instead, the model continues adapting progressively, using the weights after each adaptation as the starting point for the next sampling variation \(\mathcal{P}_v\), thereby adapting iteratively.

\begin{figure}[!b]
  \centering
  \begin{minipage}[t]{0.48\textwidth}
    \includegraphics[width=\linewidth]{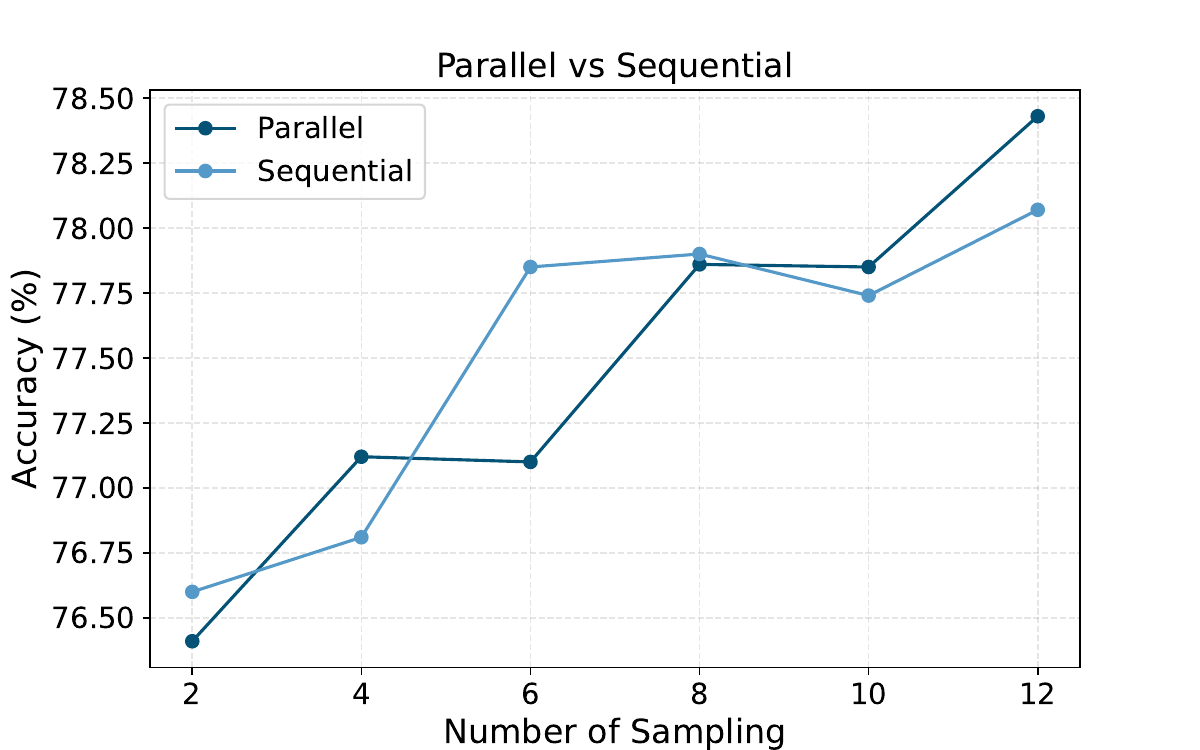} % Set width to \linewidth or less to ensure it fits within the column and is centered
    \caption{Impact of Parallel vs Sequential on Accuracy}
    %Point-MAE, Point-MAE+\gls*{geomask3d}\textsuperscript{*} and Point-MAE+\gls*{geomask3d}
    \label{fig:parallel_sequential}
  \end{minipage}\hfill
  \vspace{8pt}
  \begin{minipage}[t]{0.48\textwidth} % Adjust width to your preference
    \includegraphics[width=\linewidth]%{graphics/POINTMAE_NEW_BMVC2024_size_18__OBJBG_70_3_models_fontsize.pdf}
    {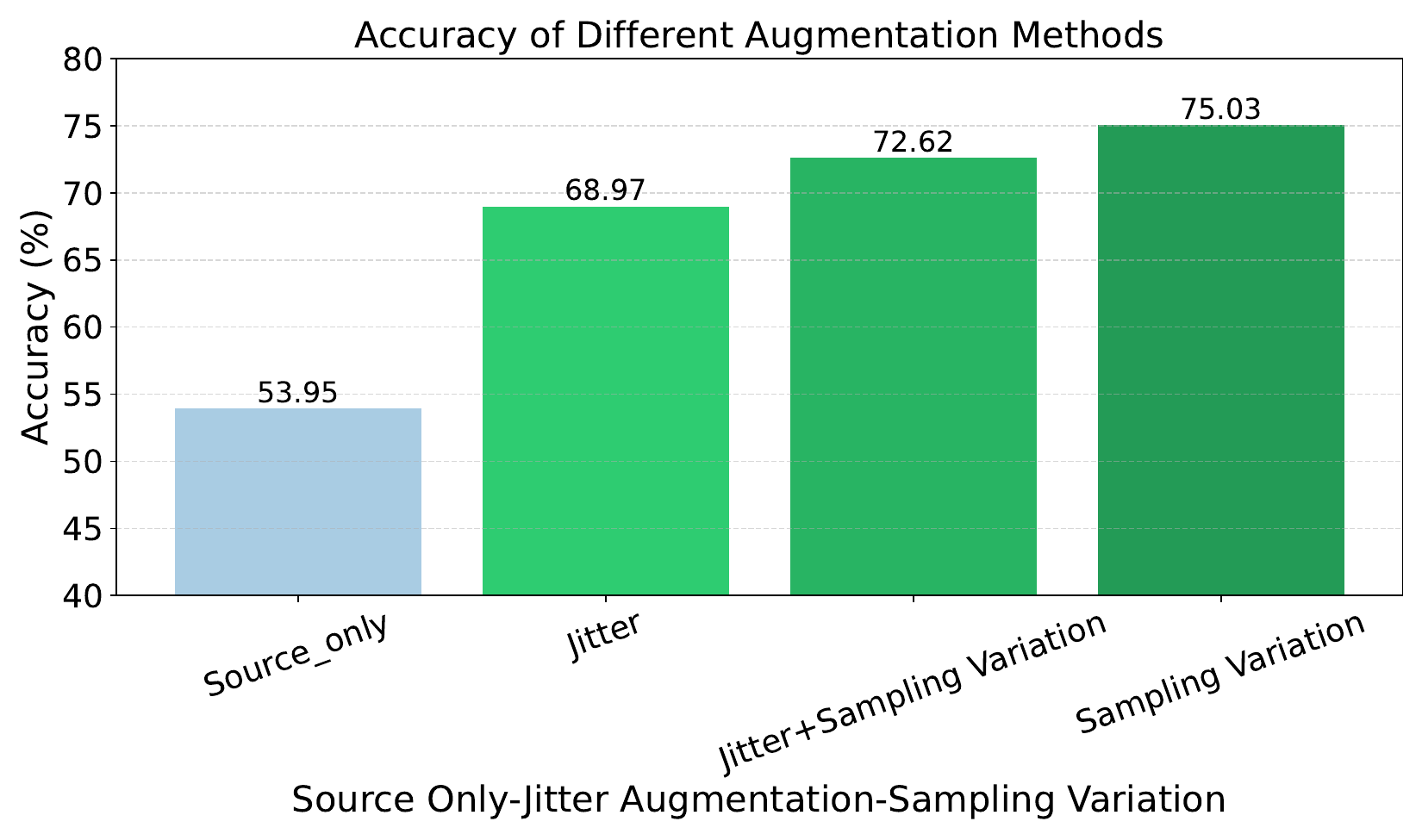}
    \caption{Comparison between Sampling Variation and Different Augmentations}
    \label{fig:aug}
  \end{minipage}
\end{figure}

% As seen in Figure~\ref{fig:parallel_sequential}, the sequential mode offers similar performance as the $N^V$ increases. However, since we aim for faster model adaptation, we select the parallel mode, as the model can be adapted in parallel with each variation $\mathcal{P}_v$.
\mypar{Parallel vs Sequential WA.}
We investigated two different strategies to handle model adaptation across multiple variations:
\begin{itemize}
\item \textbf{Parallel Mode}: After adapting the model using each variation \(\mathcal{P}_V\), the model is reset to its initial state before the next adaptation begins. The weights adapted from each variation \(\theta_v\) are stored individually. The final model weights \(\theta_{\text{avg}}\) are then calculated by averaging all the adapted weights across the variations. This approach enables the model to process each variation independently, offering faster adaptation.

\item \textbf{Sequential Mode}: In this method, the model does not reset after each adaptation. Instead, the adapted model from one variation serves as the starting point for the next variation. This results in iterative adaptation, where the model progressively refines its parameters after each variation \(\mathcal{P}_V\), creating a cumulative adaptation process. The final model weights \(\theta_{\text{avg}}\) are then calculated by averaging all the adapted weights across the variations.
\end{itemize}
As shown in Figure~\ref{fig:parallel_sequential}, both modes offer similar performance as the number of variations $N^V$ increases. However, since speed and efficiency are critical for TTA, we select the parallel mode, as it allows for faster processing by adapting the model simultaneously across all variations. This experiment was conducted using the Point-MAE backbone on the ModelNet-40C dataset.

\begin{table*}[b!]
\setlength\tabcolsep{4pt}
\centering
\resizebox{0.9\textwidth}{!}{
\small
\begin{tabular}{l|l|ccccccccccccccc|c}
\toprule
&
Method & \rotatebox{60}{uni} & \rotatebox{60}{gauss} & \rotatebox{60}{backg} & \rotatebox{60}{impul} & \rotatebox{60}{upsam} & \rotatebox{60}{rbf} & \rotatebox{60}{rbf-inv} & \rotatebox{60}{den-dec} & \rotatebox{60}{dens-inc} & \rotatebox{60}{shear} & \rotatebox{60}{rot} & \rotatebox{60}{cut} & \rotatebox{60}{distort} & \rotatebox{60}{oclsion} & \rotatebox{60}{lidar} & \rotatebox{0}{Mean}\\
\midrule
\parbox[t]{4mm}{\multirow{6}{*}{\rotatebox[origin=c]{90}{CurveNet}}} &
Source-Only & 67.3&	77.1&	7.6	& 47.6&	70.1	&78.6&	80.6&	79.2&	88.1	&77.0	&68.8&	78.6&	77.6&	35.5&	26.5&   64.0\\
&
SHOT \cite{liang2020we}   & 75.5&	78.3&	22.4	&61.1&	68.7	&72.9	&69.1	&62.3	&64.7&	39.2	&31.0&	30.6&	27.1	&10.7&	8.0&	48.1       \\
&
DUA \cite{mirza2022norm} &  81.5&	84.3&	27.5	&71.1	&81.3&	82.6&	84.5&	85.5&	89.0&	82.1	&76.9&	85.2	&81.7	&46.6&	45.8&	73.7          \\

& 
PL \cite{lee2013pseudo} &  79.5&	84.0&	29.5&	72.6	&82.7&	82.0&	83.1&	85.9&	88.7&	81.2&	78.9&	85.3&	81.6&	52.8&	52.5&	74.7\\
\cmidrule{2-18}
& 
TENT \cite{wang2020tent} &  80.9&	84.9&	29.0&	73.9&	83.8&	83.1&	\textbf{85.5}&	85.2&	\textbf{89.3}&	\textbf{83.0}&	79.8&	\textbf{85.8}&	\textbf{83.6}&	50.2&	51.0&	75.3\\
														
&
\ccol Ours & \ccol \textbf{80.9}	& \ccol \textbf{85.6}	&\ccol \textbf{30.0}	&\ccol \textbf{74.7}  &\ccol \textbf{83.9}	&\ccol \textbf{83.2}&	\ccol 84.3&	\ccol \textbf{86.1}&	\ccol 88.7	&\ccol 82.8	&\ccol \textbf{81.2}	&\ccol 85.7	&\ccol 82.7&	\ccol \textbf{56.3}&	\ccol \textbf{56.4}& \ccol \textbf{76.2}   \\

% LAME & DGCNN & 23.4 & 15.3 & 4.0 & 4.5 & 32.2 & 6.6 & 8.8 & 40.1 & 65.2 & 3.5 & 2.5 & 31.48 & 7.5 & 4.0 & 4.0 & 16.8 \\
\bottomrule
\end{tabular}}
\caption{Top-1 Classification Accuracy (\%) for all distribution shifts in the ModelNet-40C dataset.}
\label{tab:modelnet40c_2}
\end{table*}

\begin{figure*}[t!]
    \centering
    \includegraphics[width=1\textwidth]{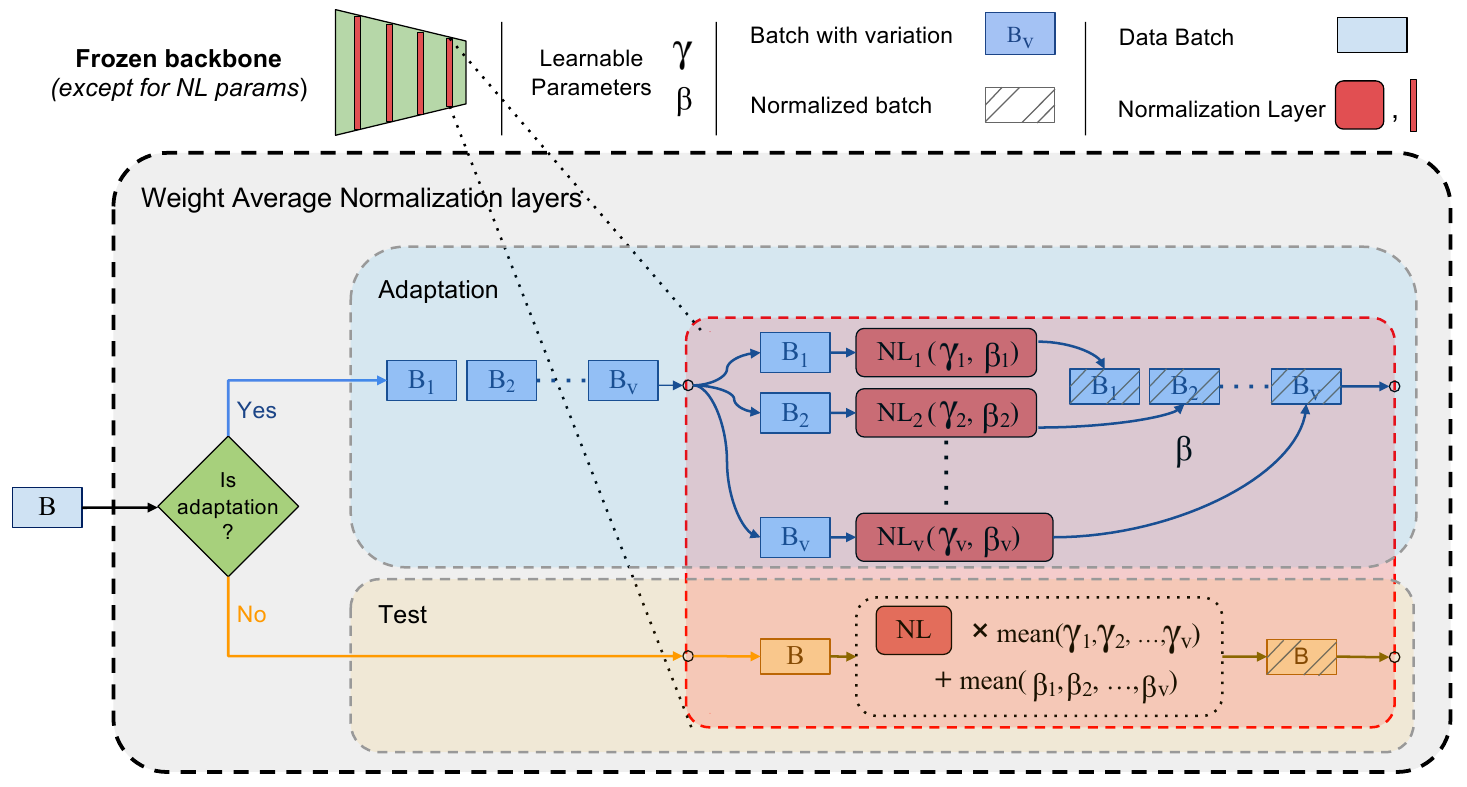}
    \caption{Detailed diagram of our method's Parallel mode.}
    \label{fig:imp_details}
\end{figure*}

\mypar{Integration of Jitter and Sampling Variation.}
In this experiment, we investigate the effect of combining jitter augmentation with the sampling variation as a new strategy \( \mathcal{P} \) to generate model diversity in our method. As seen in Figure~\ref{fig:aug}, jitter is selected for this combination because it shows the best performance among other augmentation techniques (as noted in Figure 3 of the main paper). However, while combining jitter with sampling variation yields better results compared to using jitter alone, it does not surpass the performance of our method when using sampling variation exclusively. %This highlights the effectiveness of sampling variation combined with our weight averaging technique, demonstrating its clear advantage in enhancing model robustness.

\mypar{Impact of Batch Normalization and Layer Normalization.}
In Table~\ref{tab:bn_ln_results}, we investigate the effect of updating Batch Normalization (BN) layers only versus updating both Batch Normalization (BN) and Layer Normalization (LN) layers during test-time adaptation. The results demonstrate that updating only BN layers significantly improves performance over the Source-Only baseline. Furthermore, updating both BN and LN layers leads to a slight but consistent improvement across most corruptions, resulting in a higher mean accuracy (75.0\%) compared to updating BN layers alone (74.9\%). The experiment was conducted with a batch size of 128 and 5 iterations, using PointMAE as the backbone. The dataset used was ModelNet40-C, and weight averaging was performed in parallel mode.

\mypar{Evaluation on the CurveNet Backbone.}
In order to further assess the robustness and generalizability of our method, we conducted additional experiments using a different backbone architecture, CurveNet, on the ModelNet-40C dataset. The results are summarized in Table~\ref{tab:modelnet40c_2}. As can be seen, our method demonstrates consistent improvements over baseline approaches, achieving a mean accuracy of \textbf{76.2\%}, which is notably higher than TENT’s accuracy of \textbf{75.3\%}. 
The improvements are particularly significant in corruptions like \textit{occlusion} (56.3\%), and \textit{lidar} (56.4\%), where our method consistently outperforms the other approaches.

\mypar{Efficient Parallel Implementation.}
Figure~\ref{fig:imp_details} illustrates the detailed implementation of our method in parallel mode. When adapting only the normalization layers, we handle \( N^V \) variations \( \mathcal{P}_v \) in parallel. For each sampling variation \( \mathcal{P}_v \), our method adapts the corresponding normalization layers independently. This means that the weights of the rest of the network (the majority) are shared across variations, reducing the memory overhead significantly. As shown in Figure~\ref{fig:imp_details}, we construct a ``Weight Average Normalization Layer,'' which comprises the \( N^V \) individual normalization layers.

During adaptation, all the variations are processed through their respective normalization layers. After adaptation, the normalization layer parameters \( \gamma \) and \( \beta \) are then averaged to produce the final set of normalized parameters. With this technique, we avoid saving or reloading the backbone weights for each variation, which leads to memory efficiency. For example, in the PointNet backbone, the normalization layers constitute only 0.3\% of the total network parameters. Hence, by adapting only these layers, we reduce the memory resource overhead to a mere 1.8\% when using $N^V\!=\!6$, compared to the $500\%$ memory overhead of the naive implementation.

\clearpage 
\bibliographystyle{unsrt}
\bibliography{main.bib}

\end{document}